\documentclass[sigconf]{acmart}
\AtBeginDocument{%
  \providecommand\BibTeX{{%
    \normalfont B\kern-0.5em{\scshape i\kern-0.25em b}\kern-0.8em\TeX}}}

\copyrightyear{2021} 
\acmYear{2021} 
\setcopyright{acmlicensed}\acmConference[KDD '21]{Proceedings of the 27th ACM SIGKDD Conference on Knowledge Discovery and Data Mining}{August 14--18, 2021}{Virtual Event, Singapore}
\acmBooktitle{Proceedings of the 27th ACM SIGKDD Conference on Knowledge Discovery and Data Mining (KDD '21), August 14--18, 2021, Virtual Event, Singapore}
\acmPrice{15.00}
\acmDOI{10.1145/3447548.3467319}
\acmISBN{978-1-4503-8332-5/21/08}

\usepackage{array,enumitem,multirow,caption,graphicx,subcaption,multicol,lipsum,float}
\usepackage{amsmath, amsfonts, amsthm}
\usepackage[ruled,vlined,linesnumbered]{algorithm2e}
\usepackage{kotex} 
\usepackage{makecell}
\usepackage{color, colortbl}
\usepackage{color}
\usepackage{array,enumitem,multirow,caption,graphicx,subcaption,multicol,lipsum,float,adjustbox}
\usepackage{amsmath, amsfonts, amsthm, balance}
\usepackage[colorinlistoftodos,textsize=tiny]{todonotes}
\newcolumntype{C}[1]{>{\centering\let\newline\\\arraybackslash\hspace{0pt}}m{#1}}
\newcolumntype{Z}{>{\centering\arraybackslash}m{0.062\linewidth}}
\usepackage[compatible]{algpseudocode} 
\renewcommand{\algorithmiccomment}[1]{\bgroup\hfill\small//~#1\egroup}

\begin{document}
\fancyhead{}
\title{Topology Distillation for Recommender~System}

\author{SeongKu Kang, Junyoung Hwang, Wonbin Kweon, Hwanjo Yu\*}
\affiliation{%
   \institution{Pohang University of Science and Technology (POSTECH), South Korea}
   \country{}
   \{seongku, jyhwang, kwb4453, hwanjoyu\}@postech.ac.kr
}
\authornotemark[0]
\authornote{Corresponding Author}

\begin{abstract}
Recommender Systems (RS) have employed knowledge distillation which is a model compression technique training a compact student model with the knowledge transferred from a pre-trained large teacher model. 
Recent work has shown that transferring knowledge from the teacher's intermediate layer significantly improves the recommendation quality of the student.
However, they transfer the knowledge of individual representation point-wise and thus have a limitation in that primary information of RS lies in the relations in the representation space.
This paper proposes a new topology distillation approach that guides the student by transferring the topological structure built upon the relations in the teacher space.
We first observe that simply making the student learn the whole topological structure is not always effective and even degrades the student's performance.
We demonstrate that because the capacity of the student is highly limited compared to that of the teacher, learning the whole topological structure is daunting for the student.
To address this issue, we propose a novel method named Hierarchical Topology Distillation (\proposed) which distills the topology hierarchically to cope with the large capacity gap.
Our extensive experiments on real-world datasets show that the proposed method significantly outperforms the state-of-the-art competitors. 
We also provide in-depth analyses to ascertain the benefit of distilling the topology for RS.

\end{abstract}

\newcommand{\naive}{FTD\xspace}
\newcommand{\proposed}{HTD\xspace}

\maketitle

\section{Introduction}
\label{sec:introduction}
The size of recommender systems (RS) has kept increasing, as they have employed deep and sophisticated model architectures to better understand the complex nature of user-item interactions \cite{RD, CD, DERRD,  GCN_distill}.
A large model with many learning parameters has a high capacity and therefore generally achieves higher recommendation accuracy.
However, it also requires high computational costs, which results in high inference latency.
For this reason, it is challenging to adopt such a large model to the real-time platform \cite{RD, CD, GCN_distill, DERRD}.

To tackle this problem, \textit{Knowledge Distillation} (KD) has been adopted to RS \cite{RD, CD, DERRD, zhu2020ensembled, BD, GCN_distill}.
KD is a model-independent strategy to improve the performance of a compact model (i.e., student) by transferring the knowledge from a pre-trained large model (i.e., teacher).
The distillation is conducted in two steps.
The teacher is first trained with the training set, and the student is trained with help from the teacher along with the training set.
The student model, which is a compact model, is used in the inference time.
During the distillation, the teacher can provide additional supervision that is not explicitly revealed from the training set.
As a result, the student trained with KD shows better prediction performance than the student trained only with the training set.
Also, it has low inference latency due to its small size.

\begin{figure*}[t]
\centering
\captionsetup[subfigure]{justification=centering}
\begin{subfigure}[t]{0.33\linewidth}
    \includegraphics[width=\linewidth]{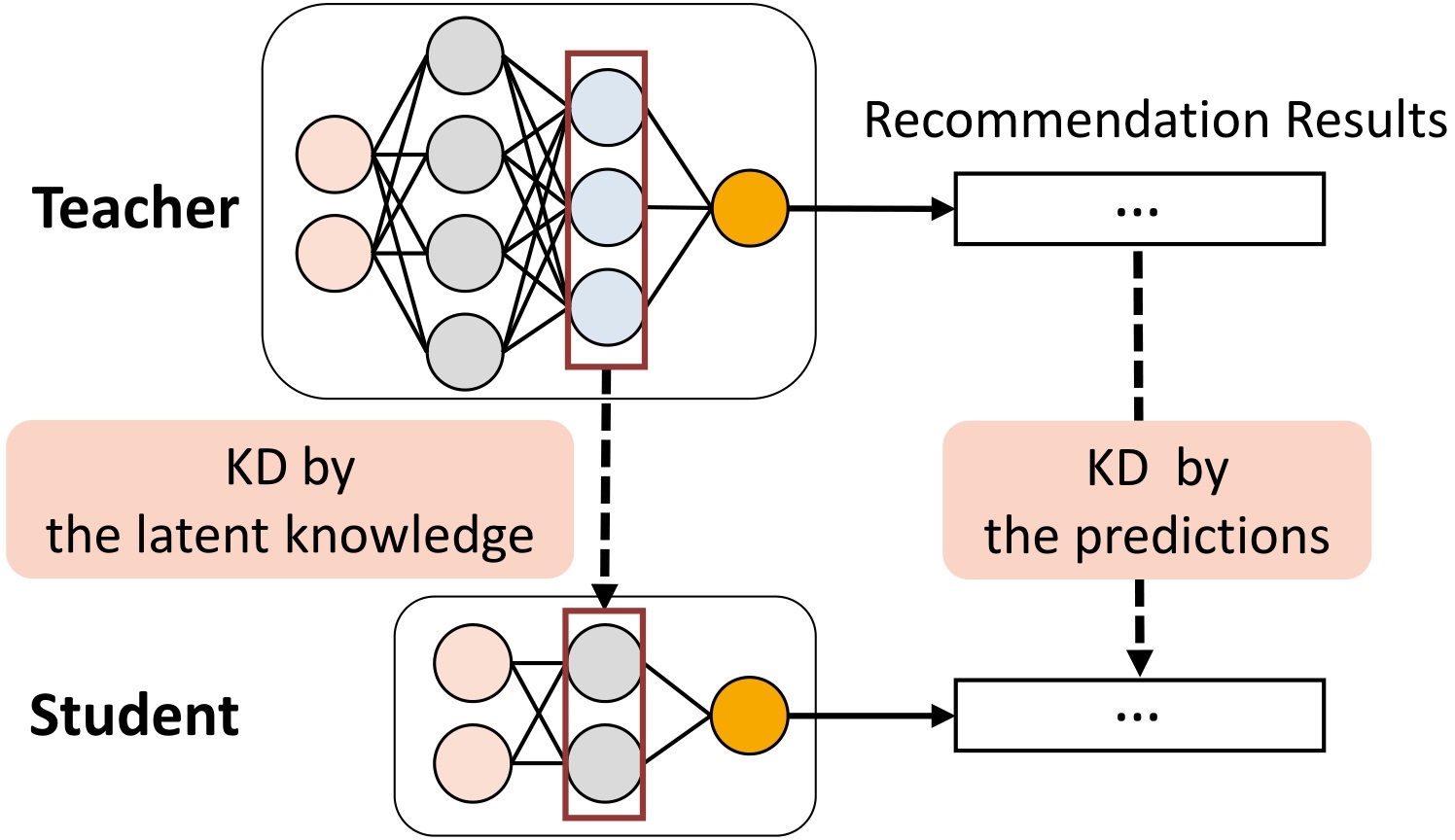}
    \subcaption{}
\end{subfigure}
\hspace{10pt}
\begin{subfigure}[t]{0.16\linewidth}
    \includegraphics[width=\linewidth]{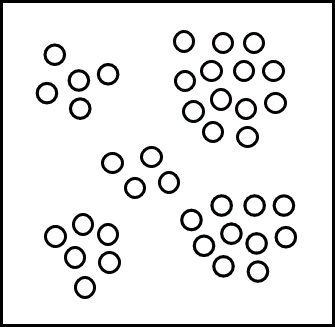}
    \subcaption{\footnotesize Hint Regression}
\end{subfigure}
\begin{subfigure}[t]{0.16\linewidth}
    \includegraphics[width=\linewidth]{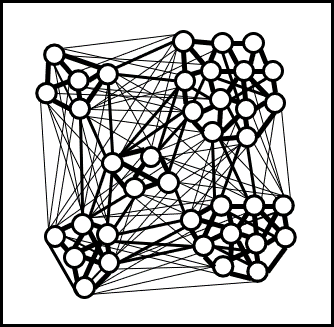}
    \subcaption{\footnotesize FTD (Sec.3.2)}
\end{subfigure}
\begin{subfigure}[t]{0.16\linewidth}
    \includegraphics[width=\linewidth]{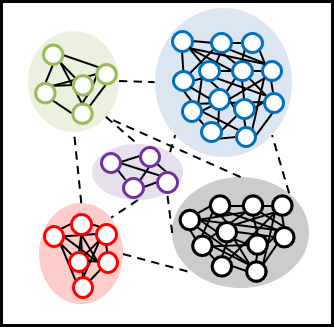}
    \subcaption{\footnotesize HTD~(Sec.3.3)}
\end{subfigure}
\caption{(a) The overview of KD in RS. (b-d) The conceptual illustrations of the latent knowledge that each method transfers from the teacher's representation space. Each point corresponds to a representation of each entity. 
(b) transfers the information of each entity to the student point-wise.
However, our approach (c-d) transfers the relations among the entities.
FTD/HTD refers to Full/Hierarchical topology distillation, and the solid/dotted line denotes the entity/group-level topology, respectively.}
\label{fig:intro}
\vspace{-0.4cm}
\end{figure*}

Most existing KD methods for RS transfer the knowledge from the teacher's predictions \cite{DERRD, RD, CD, GCN_distill, BD} (Figure \ref{fig:intro}a).
They basically enforce the student to imitate the teacher's recommendation results, providing guidance to the predictions of the student.
There is another recent approach that transfers the \emph{latent} knowledge from the teacher's intermediate layer \cite{DERRD, zhu2020ensembled}, pointing out that the predictions incompletely reveal the teacher’s knowledge and the intermediate representation can additionally provide a detailed explanation on the final prediction of the teacher.
They adopt \textit{hint regression} \cite{FitNet} that makes the student's representation approximate the teacher's representation via a few regression layers.
This enables the student to get compressed information on each entity (e.g., user and item) (Figure \ref{fig:intro}b) that can restore more detailed preference information in the teacher \cite{DERRD}.

However, the existing hint regression-based methods focus on distilling the individual representation of each entity, disregarding the \emph{relations} of the representations.
In RS, each entity is better understood by its relations to the other entities rather than by its individual representation.
For instance, a user's preference is represented in relation to (and in contrast with) other users and items.
Also, the student can take advantage of the space, where the relations found by the teacher are well preserved, in finding more accurate ranking orders among the entities and thereby improving the recommendation performance.

This paper proposes a new distillation approach that effectively transfers the relational knowledge existing in the teacher's representation space.
A natural question is how to define the relational knowledge and distill it to the student.
We build a \emph{topological structure} that represents the relations in the teacher space based on the similarity information, then utilize it to guide the learning of the student via distillation.
Specifically, we train the student with the distillation loss that preserves the teacher's topological structure in its representation space along with the original loss function.
Trained with the topology distillation, the student can better preserve the relations in the teacher space, which not only improves the recommendation performance but also better captures the semantic of entities (reported in Section 4.3).

However, we observe that simply making the student learn all the topology information (Figure \ref{fig:intro}c) is not always effective and sometimes even degrades the student's recommendation performance (reported in Section 4.2).
This phenomenon is explained by the huge capacity gap between the student and the teacher;
the capacity of the student is highly limited compared to that of the teacher, and learning all the topological structure in the teacher space is often daunting for the student.
To address this issue, we propose a method named Hierarchical Topology Distillation (\proposed) which effectively transfers the vast teacher's knowledge to the student with limited capacity.
\proposed represents the topology hierarchically and transfers the knowledge in multi-levels using the hierarchy (Figure \ref{fig:intro}d).

Specifically, \proposed adaptively finds \textit{preference groups} of entities such that the entities within each group share similar preferences.
Then, the topology is hierarchically structured in group-level and entity-level.
The \emph{group-level topology} represents the summarized relations across the groups, providing an overview of the whole topology.
The \emph{entity-level topology} represents the relations of entities belonging to the same group.
This provides a fine-grained view on important relations among the entities having similar preferences, which directly affects the top-$N$ recommendation performance.
By compressing the complex individual relations across the groups, \proposed relaxes the daunting supervision and enables the student to better focus on the important relations.
In summary, the key contributions of our work are as follows:
\begin{itemize}[leftmargin=*]
    \item We address the necessity of transferring the relational knowledge from the teacher representation space and develop a general topology distillation approach for RS.
    
    \item We develop a new topology distillation method, named FTD, designed to guide the student by transferring the full topological structure built upon the relations in the teacher space.  
    
    \item We propose a novel topology distillation method, named~\proposed, designed to effectively transfer the vast relational knowledge to the student considering the huge capacity gap.
    
    \item We validate the superiority of the proposed approach by extensive experiments. 
    We also provide in-depth analyses to verify the benefit of distilling the topological structure.
\end{itemize}

\section{Related Work}
\label{sec:relatedwork}
\noindent
\textbf{Knowledge Distillation.}
Knowledge distillation (KD) is a model-independent strategy that accelerates the training of a student model with the knowledge transferred from a pre-trained teacher model.
Most KD methods have mainly focused on the image classification task.
An early work \cite{hinton2015distilling} matches the class distributions (i.e., the softmax output) of the teacher and the student.
The class distribution has richer information (e.g., inter-class correlation) than the one-hot class label, which improves learning of the student model.
Pointing out that utilizing the predictions alone is insufficient because meaningful intermediate information is ignored, subsequent methods \cite{FitNet, yim2017gift, AT, liu2019structured, tung2019similarity} have distilled knowledge from the teacher's intermediate layer along with the predictions.
\cite{FitNet} proposes ``hint regression'' that matches the intermediate representations.
Subsequently, \cite{yim2017gift} matches the gram matrices of the representations, \cite{AT} matches the attention maps from the networks, and \cite{tung2019similarity, li2020local} match the similarities on activation maps of the convolutional layer.

\vspace{0.05cm}
\noindent
\textbf{Reducing inference latency of RS.}
As the size of RS is continuously increasing, various approaches have been proposed for reducing the model size and inference latency.
Several methods \cite{hash1, hash2, DCF} have utilized the binary representations of users and items.
With the discretized representations, the search costs can be considerably reduced via the hash technique.
However, due to their restricted capability, the loss of recommendation accuracy is inevitable \cite{CD, DERRD, GCN_distill}.
Also, various computational acceleration techniques have been successfully adopted to reduce the search costs.
In specific, order-preserving transformations \cite{tree_RS}, pruning and compression techniques \cite{pruning_RS2_inner_only, pruning_RS, compression1}, tree-based data structures \cite{KDtree, tree_RS}, and approximated nearest-neighbor search \cite{LSH_inner_product} have been employed to reduce the inference latency.
However, they have limitations in that the techniques are only applicable to specific models or easy to fall into a local optimum because of the local search~\cite{DERRD, CD}.

\vspace{0.05cm}
\noindent
\textbf{Knowledge Distillation for RS.}
KD, which is the model-agnostic strategy, has been widely adopted in RS.
Similar to the progress on computer vision, the existing methods are categorized into two groups (Figure 1a):
(1) the methods distilling knowledge from the predictions, (2) the methods distilling the latent knowledge from the intermediate layer.
Note that the two groups of methods can be utilized together to fully improve the student \cite{DERRD}.

\noindent
{\textbf{(1) KD by the predictions.}}
Motivated by \cite{hinton2015distilling} that matches the class distributions, most existing methods \cite{RD, CD, DERRD, GCN_distill, BD} have focused on matching the predictions (i.e., recommendation results) from the teacher and the student.
The teacher's predictions convey additional information about the subtle difference among the items, helping the student generalize better than directly learning from binary labels \cite{CD}.
This research direction focuses on designing a method effectively utilizing the teacher's predictions.
First, \cite{RD, CD} distill the knowledge of the items with high scores in the teacher’s predictions.
Since a user is interested in only a few items, distilling knowledge of a few top-ranked items is effective to discover the user's preferable items \cite{RD}.
Most recently, \cite{BD} utilizes rank-discrepancy information between the predictions from the teacher and the student.
Specifically, \cite{BD} focuses on distilling the knowledge of the items ranked highly by the teacher but ranked lowly by the student.
On the one hand, \cite{DERRD, GCN_distill} focus on distilling ranking order information from the teacher's predictions.
Concretely, they adopt listwise learning \cite{xia2008list-wise} and train the student to follow the items' ranking orders predicted by the teacher.

\noindent
{\textbf{(2) KD by the latent knowledge.}}
Pointing out that the predictions incompletely reveal the teacher's knowledge, a few methods \cite{DERRD, zhu2020ensembled}\footnote{\cite{DERRD} proposes two KD methods: one by prediction and the other by latent knowledge.} have focused on distilling latent knowledge from the teacher's intermediate layer.
The existing methods are based on \textit{hint regression} \cite{FitNet} proposed in computer vision.
Let $h^t: \mathcal{X} \rightarrow \mathbb{R}^{d^t}$ denote a mapping function from the input feature space to the representation space of the teacher (i.e., the teacher nested function up to the intermediate layer).
Similarly, let $h^s: \mathcal{X} \rightarrow \mathbb{R}^{d^s}$ denote a mapping function to the representation space of the student.
Also, let $\mathbf{e}_i^t = h^t(\mathbf{x}_i)$ and $\mathbf{e}_i^s = h^s(\mathbf{x}_i)$ denote the representations of entity $i$ from the two spaces\footnote{We use the term `entity' to denote the subject of each representation.}, where $\mathbf{x}_i$ is entity $i$'s input feature.
The hint regression makes $\mathbf{e}_i^s$ approximate $\mathbf{e}_i^t$ as follows:
\begin{equation}
    \mathcal{L}_{Hint} = \lVert \mathbf{e}_i^t -  f(\mathbf{e}_i^s) \rVert^2_2
\end{equation}
where $f:\mathbb{R}^{d^s} \rightarrow \mathbb{R}^{d^t}$ is a small network to bridge the different dimensions ($d^s << d^t$).
By minimizing $\mathcal{L}_{Hint}$, parameters in the student (i.e., $h^s$) and $f$ are updated.
Also, it is jointly minimized with the base model (i.e., $\mathcal{L}_{B a s e} + \lambda \mathcal{L}_{H i n t}$) which can be any existing recommender.
The hint regression enables $\mathbf{e}^s$ to capture compressed information that can restore detailed information in $\mathbf{e}^t$ \cite{zhu2020ensembled, DERRD}.
\cite{zhu2020ensembled} adopts this original hint regression~to~improve~the~student.

The most recent work DE \cite{DERRD} further elaborates this approach for RS.
DE argues that using a single network ($f$) makes the knowledge of entities having dissimilar preferences get mixed, and this degrades the quality of distillation.
Its main idea is that the knowledge of entities having similar preferences should be distilled without being mixed with that of entities having dissimilar preferences. 
To this end, DE clusters the representations into $K$ groups based on the teacher’s knowledge and distills the representations in each group via a separate network $f_k$.
Let $\mathbf{z}_{i}$ be a $K$-dimensional one-hot vector whose element $z_{ik}=1$ if entity $i$ belongs to the corresponding $k$-th group.
For each entity $i$, DE loss is defined as follows:
\begin{equation}
  \begin{aligned}
    \mathcal{L}_{DE} = \lVert \mathbf{e}_i^t -  \sum_{k=1}^K z_{ik} f_k(\mathbf{e}_i^s) \rVert^2_2
\end{aligned}
\end{equation} 
The one-hot vector is sampled from a categorical distribution with class probabilities $\boldsymbol{\alpha}_i = v(\mathbf{e}^{t}_i)$, i.e., $\mathbf{z}_{i} \sim \text{Categorical}_{K} (\boldsymbol{\alpha}_i)$, where $v: \mathbb{R}^{d^t} \rightarrow \mathbb{R}^{K}$ is a small network with Softmax output.
The sampling process is approximated by Gumbel-Softmax \cite{GumbelSoftmax} and trained via backpropagation in an end-to-end manner.
In sum, the representations belonging to the same group share similar preferences and are distilled via the same network without being mixed with the representations belonging to the different groups \cite{DERRD}.

The existing methods \cite{DERRD, zhu2020ensembled} based on the hint regression distill the knowledge of individual entity without consideration of how the entities are related in the representation space. 
Considering a user’s preference is represented in relation to (and in contrast with) other users and items, each entity is better understood by its relations to the other entities rather than by its individual representation.
Also, the student can take advantage of the space, where the relations found by the teacher are well preserved, in finding more accurate ranking orders among the entities and thereby improving the recommendation performance.
In this work, we propose a new distillation approach for RS that directly distills the relational knowledge from the teacher's representation space.

\section{Methodologies}
\label{sec:method}
We first provide an overview of the proposed approach (Section 3.1).
Before we describe the final solution, we explain a naive method for incorporating the relational knowledge in the distillation process (Section 3.2).
Then, we shed light on the drawbacks of the method when applying it for KD.
Motivated by the analysis, we present a new method, named \proposed, which distills the relational knowledge in multi-levels to effectively cope with a large capacity gap between the teacher and the student (Section 3.3).
The pseudocodes of the proposed methods are provided in the appendix.

\subsection{Overview of Topology Distillation}
The proposed topology distillation approach guides the learning of the student by the topological structure built upon the relational knowledge in the teacher representation space.
The relational knowledge refers to all the information on how the representations are correlated in the space;
those sharing similar preferences are strongly correlated, whereas those with different preferences are weakly correlated.
We build a (weighted) topology of a graph where the nodes are the representations and the edges encode the relatedness of the representations.
Then, we distill the relational knowledge by making the student preserve the teacher's topological structure in its representation space.
With the proposed approach, the student is trained by minimizing the following loss:
\begin{equation}
    \mathcal{L} = \mathcal{L}_{Base} + \lambda_{TD} \mathcal{L}_{TD}
\end{equation}
where $\lambda_{TD}$ is a hyperparameter controlling the effects of topology distillation.
The base model can be any existing recommender, and $\mathcal{L}_{Base}$ corresponds to its loss function.
$\mathcal{L}_{TD}$ is defined on the topology of the representations in the same batch used for the base model training.

\subsection{Full Topology Distillation (FTD)}
As a straightforward method, we distill the knowledge of the entire relations in the teacher space.
Given a batch, we first generate a fully connected graph in the teacher representation space.
The graph is characterized by the adjacency matrix $\mathbf{A}^t \in \mathbb{R}^{b \times b}$ where $b$ is the number of representations in the batch. 
Each element $a^t_{ij}$ is the weight of an edge between entities $i$ and $j$ representing their similarity and is parameterized as follows:
\begin{align}
    a^t_{ij}=\rho(\mathbf{e}^t_i,\mathbf{e}^t_j)
\end{align}
where $\rho(\cdot,\cdot)$ is a similarity score such as the cosine similarity or negative Euclidean distance, in this work we use the former.
Analogously, we generate a graph, characterized by the adjacency matrix $\mathbf{A}^s\in \mathbb{R}^{b \times b}$, in the student representation space, i.e., $a^s_{ij}=\rho(\mathbf{e}^s_i,\mathbf{e}^s_j)$.

After obtaining the topological structures $\mathbf{A}^t$ and $\mathbf{A}^s$ from the representation space of the teacher and student, respectively, we train the student to preserve the topology discovered by the teacher by the topology-preserving distillation loss as follows:
\begin{align}
    \mathcal{L}_{F T D} = \text{Dist}(\mathbf{A}^t, \mathbf{A}^s) =  \lVert \mathbf{A}^t - \mathbf{A}^s \rVert^2_F,
\end{align}
where $\text{Dist}(\cdot, \cdot)$ is the distance between the topological structures, in this work, we compute it with the Frobenius norm.
By minimizing $\mathcal{L}_{FTD}$, parameters in the student are updated.
As this method utilizes the full topology as supervision to guide the student, we call it Full Topology Distillation (FTD).
Substituting the distillation loss $\mathcal{L}_{TD}$ in Equation 3 derives the final loss for training the student.

\noindent
\textbf{Issues:}
Although FTD directly transfers the relational knowledge which is ignored in the previous work, it still has some clear drawbacks.
Because the student has a very limited capacity compared to the teacher, it is often daunting for the student to learn all the relational knowledge in the teacher.
Indeed, we observe that sometimes FTD even hinders the learning of the student and degrades the recommendation performance (reported in Section 4.2).
Therefore, the relational knowledge should be distilled with consideration of the huge capacity gap.

\subsection{Hierarchical Topology Distillation (HTD)}
Our key idea to tackle the issues is to decompose the whole topology hierarchically so as to be effectively transferred to the student.
We argue that the student should focus on learning the relations among the strongly correlated entities that share similar preferences and accordingly have a direct impact on top-$N$ recommendation performance.
To this end, we summarize the numerous relations among the weakly correlated entities, enabling the student to better focus on the important relations.

During the training, \proposed adaptively finds \emph{preference groups} of strongly correlated entities.
Then, the topology is hierarchically structured in group-level and entity-level:
1) \emph{group-level topology} includes the summarized relations across the groups, providing the overview of the entire topology.
2) \emph{entity-level topology} includes the relations of entities belonging to the same group.
This provides fine-grained supervision on important relations of the entities having similar preferences.
By compressing the complex individual relations across the groups, \proposed relaxes the daunting supervision, effectively distills the relational knowledge to the student.

\subsubsection{\textbf{Preference Group Assignment}}
To find the groups of entities having a similar preference in an end-to-end manner considering both the teacher and the student, we borrow the idea of DE \cite{DERRD}.
Formally, let there exist $K$ preference groups in the teacher space.
We use a small network $v: \mathbb{R}^{d^t} \rightarrow \mathbb{R}^{K}$ with Softmax output to compute the assignment probability vector $\boldsymbol{\alpha}_i \in \mathbb{R}^{K}$ for each entity $i$ as follows:
\begin{equation}
    \boldsymbol{\alpha}_{i} = v( \mathbf{e}^{t}_i),
\end{equation}
where each element $\alpha_{ik}$ encodes the probability of the entity $i$ to be assigned to $k$-th preference group.
Let $\mathbf{z}_{i}$ be a $K$-dimensional one-hot assignment vector whose element $z_{ik}=1$ if entity $i$ belongs to the corresponding  $k$-th group.
We assign a group for each entity by sampling the assignment vector from a categorical distribution parameterized by $\{\alpha_{ik}\}$ i.e., $p(z_{ik}=1 \mid v, \mathbf{e}^t_i)=\alpha_{ik}$.
To make the sampling process differentiable, we adopt Gumbel-Softmax \cite{GumbelSoftmax} which is a continuous distribution on the simplex that can approximate samples from a categorical distribution.
\begin{equation}
\begin{aligned}
z_{ik} =\frac{\exp\left(\left( \alpha_{ik} + g_k \right) / \tau\right) }{\sum_{j=1}^K \exp\left(\left( \alpha_{ij} + g_j \right) / \tau\right)} \quad \text{for} \quad k=1,...,K,
\end{aligned}
\end{equation}
where $g_j$ is the gumbel noise drawn from $\text{Gumbel}(0,1)$ distribution \cite{GumbelSoftmax} and $\tau$ is the temperature parameter.
We set a small value on $\tau$ so that samples from the Gumbel-Softmax distribution become one-hot vector \cite{GumbelSoftmax}.
This group assignment is evolved during the training via backpropagation \cite{DERRD}. 
This will be explained in Section~3.3.4.

With the assignment process, for a given batch, \proposed obtains the grouping information summarized by a $b \times K$ assignment matrix $\mathbf{Z}$ where each row corresponds to the one-hot assignment vector for each entity.
We also denote the set of entities belonging to each group by $G_{k}=\{i|z_{ik}=1\}$.
Note that the group assignment is based on the teacher's knowledge.
Based on the assignment, we decompose the topology hierarchically.

\subsubsection{\textbf{Group-level topology.}}
\proposed introduces a \textit{prototype} representing the entities in each preference group, then use it to summarize the relations across the groups.
Let $\mathbf{E}^t \in \mathbb{R}^{b \times d^t}$ and $\mathbf{E}^s \in \mathbb{R}^{b \times d^s}$ denote the representation matrix in the teacher space and the student space, respectively.
The prototypes $\mathbf{P}^t \in \mathbb{R}^{K \times d^t}$ and $\mathbf{P}^s \in \mathbb{R}^{K \times d^s}$ are defined as follows:
\begin{equation}
\begin{aligned}
    \mathbf{P}^t = \tilde{\mathbf{Z}}^\top \mathbf{E}^t \quad\text{and}\quad \mathbf{P}^s = \tilde{\mathbf{Z}}^\top \mathbf{E}^s,
\end{aligned}
\end{equation}
where $\tilde{\mathbf{Z}}$ is normalized assignment matrix by the number of entities in each group (i.e., $\tilde{\mathbf{Z}}_{[:,i]} = \mathbf{Z}_{[:,i]} / \sum_i \mathbf{Z}_{[:,i]}$).
For $\mathbf{P}$, each row $\mathbf{P}_{[k,:]}$ corresponds to the average representation for the entities belonging to each group $k$, and we use it as a prototype representing the group.

With the prototypes, we consider two design choices with different degrees of relaxation.
In the first choice, we distill the relations between the prototypes.
We build the topology characterized by the $K \times K$ matrix $\mathbf{H}^t$ which contains the relations~as:
\begin{equation}
h^t_{km} = \rho(\mathbf{P}^t_{[k,:]}, \mathbf{P}^t_{[m,:]}),
\end{equation}
where $k, m \in \{1, ..., K\}$.

In the second choice, we distill the relations between each prototype and entities belonging to the other groups.
We build the topology characterized by the $K \times b$ matrix $\mathbf{H}^t$ which contains the relations~as:
\begin{equation}
h^t_{kj} = \rho(\mathbf{P}^t_{[k,:]}, \mathbf{e}^t_j),
\end{equation}
where $k \in \{1, ..., K\}, j \in \{1, ..., b\}$.
It is worth noting that we only distill the relations across the groups (i.e., $j \notin G_k$).
Using one of the choices, we build the group-level topological structure $\mathbf{H}^t$ in the teacher space, and analogously, we build $\mathbf{H}^s$ in the student space.

The first choice puts a much higher degree of relaxation compared to the second choice.
For instance, assume that there are two groups of entities (i.e., $G_1$ and $G_2$).
Without the hierarchical approach (as done in FTD), there exists $|G_1|\times|G_2|$ relations across the groups.
With the first choice, they are summarized to a single relation between two prototypes, and with the second choice, they are summarized to $|G_1|+|G_2|$ relations.
We call the first choice as Group(P,P) and the second choice as Group(P,e) and provide results with each choice in Section 4.2.

\subsubsection{\textbf{Entity-level topology.}}
\proposed distills the full relations among the strongly correlated entities in the same group.
In the teacher space, the entity-level topology contains the following~relations:
\begin{equation}
\{ \rho(\mathbf{e}^t_i,\mathbf{e}^t_j) \mid (i, j) \in G_{k} \times G_{k} \}, \;\;\text{for} \;k \in \{1, ..., K\},
\end{equation}
and analogously, we build the entity-level topology in the student space. 
For an efficient computation on matrix form, we introduce the $b \times b$ binary indicator matrix $\mathbf{M} = \mathbf{Z} \mathbf{Z}^\top$ indicating whether each relation is contained in the topology or not.
Intuitively, each element $m_{ij}=1$ if entity $i$ and $j$ are assigned to the same group, otherwise $m_{ij}=0$.
Then, the entity-level topology is defined by $\mathbf{A}^t$ with $\mathbf{M}$ in the teacher space and also defined by $\mathbf{A}^s$ with $\mathbf{M}$ in the student space.
The distance between two topological structures is simply computed by $\lVert \mathbf{M} \odot (\mathbf{A}^t - \mathbf{A}^s) \rVert^2_F$ where  $\odot$ is the Hadamard product.

\subsubsection{\textbf{Optimization}}
\proposed guides the student with the decomposed topological structures.
The loss function is defined as follows:
\begin{equation}
\begin{aligned}
\mathcal{L}_{HTD} &= 
\gamma  \left( \lVert \mathbf{H}^t - \mathbf{H}^s \rVert^2_F + \lVert \mathbf{M} \odot (\mathbf{A}^t - \mathbf{A}^s) \rVert^2_F \right)\\
&+ (1-\gamma) \left(\sum_{i=1}^b \lVert \mathbf{e}^t_i - \sum^K_{k=1} z_{ik}  f_k(\mathbf{e}^s_i) \rVert^2_2\right),
\end{aligned}
\end{equation}
where the first term corresponds to the topology-preserving loss,
the second term corresponds to the hint regression loss adopted in DE \cite{DERRD} that makes the group assignment process differentiable.
We put a network $f_k$ for each $k$-th group, then train each network to reconstruct the representations belonging to the corresponding group, which makes the entities having strong correlations get distilled by the same network \cite{DERRD}.
$\gamma$ is a hyperparameter balancing the two terms. 
In this work, we set 0.5 to $\gamma$.
By minimizing $\mathcal{L}_{HTD}$, parameters in the student, $v$ and $f_*$ are updated.
Note that $v$ and $f_*$ are not used in the inference phase, they are only utilized for the distillation in the offline training phase.
Substituting the distillation loss $\mathcal{L}_{TD}$ in Equation 3 derives the final loss for training the student.

\begin{figure}[t]
\centering
\begin{subfigure}[t]{0.95\linewidth}
    \includegraphics[width=\linewidth]{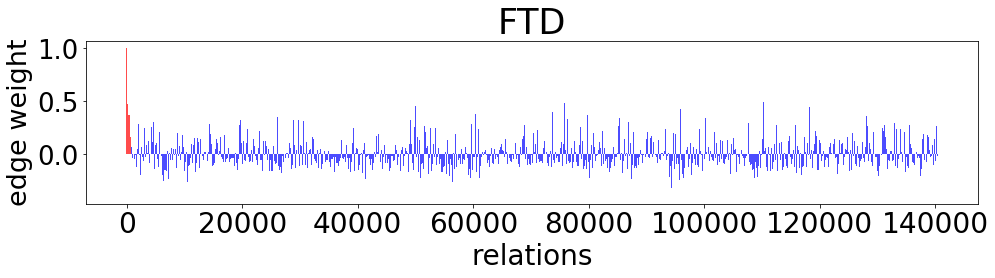}
\end{subfigure}
\begin{subfigure}[t]{0.95\linewidth}
    \includegraphics[width=\linewidth]{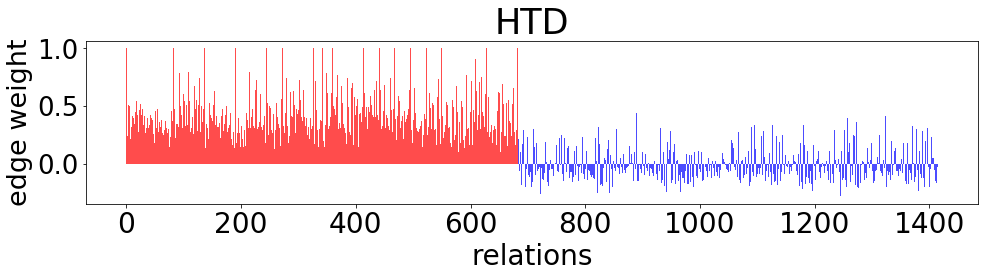}
\end{subfigure}
\caption{The relational knowledge distilled from teacher to student by FTD and HTD (with Group(P,e)). Red/Blue corresponds to relations of the entities belonging to the same/different preference group(s) (BPR on CiteULike).}
\label{fig:FTDvsHTD}
\vspace{-0.5cm}
\end{figure}

\vspace{0.1cm}
\noindent
\textbf{Effects of \proposed.}
For more intuitive understanding, we provide a visualization of the relational knowledge distilled to the student in Figure \ref{fig:FTDvsHTD}.
We randomly choose a preference group and visualize the relations \textit{w.r.t.} the entities belonging to the group.
Note that there exist the same number of intra-group relations (red) in both figures.
Without the hierarchical approach (as done in FTD), the student is forced to learn a huge number of relations with the entities belonging to the other groups (blue).
On the other hand, \proposed summarizes the numerous relations, enabling the student to better focus on learning the detailed intra-group relations, which directly affects in improving the recommendation performance.

\vspace{0.1cm}
\noindent
\textbf{Discussions on Group Assignment.}
Note that \proposed is not limited to a specific group assignment method, i.e., DE.
Any method that clusters the teacher representation space or prior knowledge of user/item groups (e.g., item category, user demographic features) can be utilized for more sophisticated topology decomposition, which further improves the effectiveness of \proposed. 
The comparison with another assignment method is provided in the appendix.

\section{Experiments}
\label{sec:experimentsetup}
We validate the proposed approach on \textbf{18} experiment settings: 2 real-world datasets × 3 base models × 3 different student model sizes (Section 4.1).
We first present comparison results with the state-of-the-art competitor and a detailed ablation study (Section 4.2).
We also provide in-depth analyses to verify the benefit of distilling the topological structure (Section 4.3).
Lastly, we provide a hyperparameter study (Section 4.4).

\subsection{Experimental Setup}
We closely follow the experiment setup of DE \cite{DERRD}.
However, for a thorough evaluation, we make two changes in the setup.
\textbf{1)} we add LightGCN \cite{he2020lightgcn}, which is the state-of-the-art top-$N$ recommendation method, as a base model.
\textbf{2)} unlike \cite{DERRD} that samples negative items for evaluation, we adopt the full-ranking evaluation which enables more rigorous evaluation.
Refer to the appendix for more detail.

\subsubsection{Datasets}
We use two real-world datasets: CiteULike and Foursquare.
CiteULike contains tag information for each item, and Foursquare contains GPS coordinates for each item.
We use the side information to evaluate the quality of representations induced by each KD method.
More details of the datasets are provided in the appendix.

\subsubsection{Base Models}
We evaluate the proposed approach on base models having different architectures and learning strategies, which are widely used for top-$N$ recommendation task.
\begin{itemize}[leftmargin=*]
    \item \textbf{BPR \cite{BPR}}: 
    A learning-to-rank model that models user-item interaction with Matrix Factorization (MF).
    \item \textbf{NeuMF \cite{NeuMF}}: A deep model that combines MF and Multi-Layer Perceptron (MLP) to learn the user-item interaction.
    \item \textbf{LightGCN \cite{he2020lightgcn}}: The state-of-the-art model which adopts simplified Graph Convolution Network (GCN) to capture the information of multi-hop neighbors.
\end{itemize}

\subsubsection{Teacher/Student}
\begin{figure}[t]
\begin{subfigure}[t]{0.95\linewidth}
    \includegraphics[width=\linewidth]{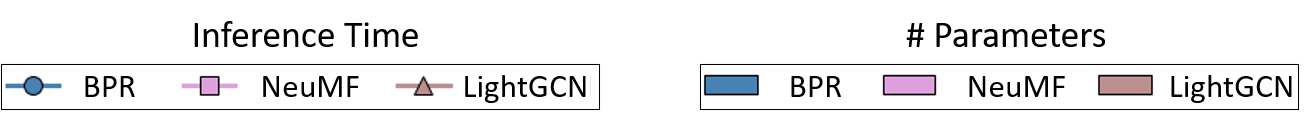}
\end{subfigure}\\
\begin{subfigure}[t]{0.45\linewidth}
    \includegraphics[width=\linewidth]{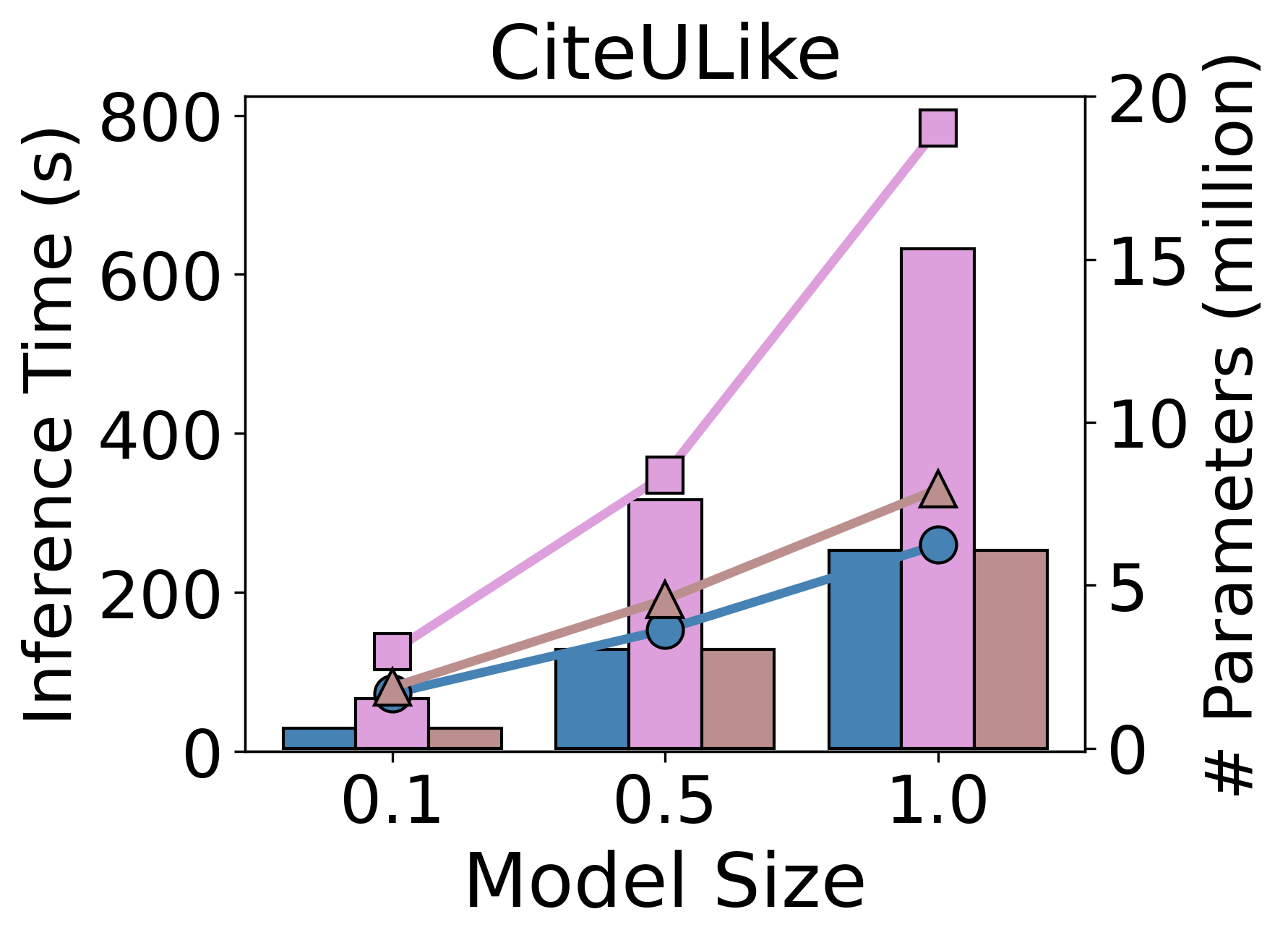}
\end{subfigure}
\hspace{0.1cm}
\begin{subfigure}[t]{0.47\linewidth}
    \includegraphics[width=\linewidth]{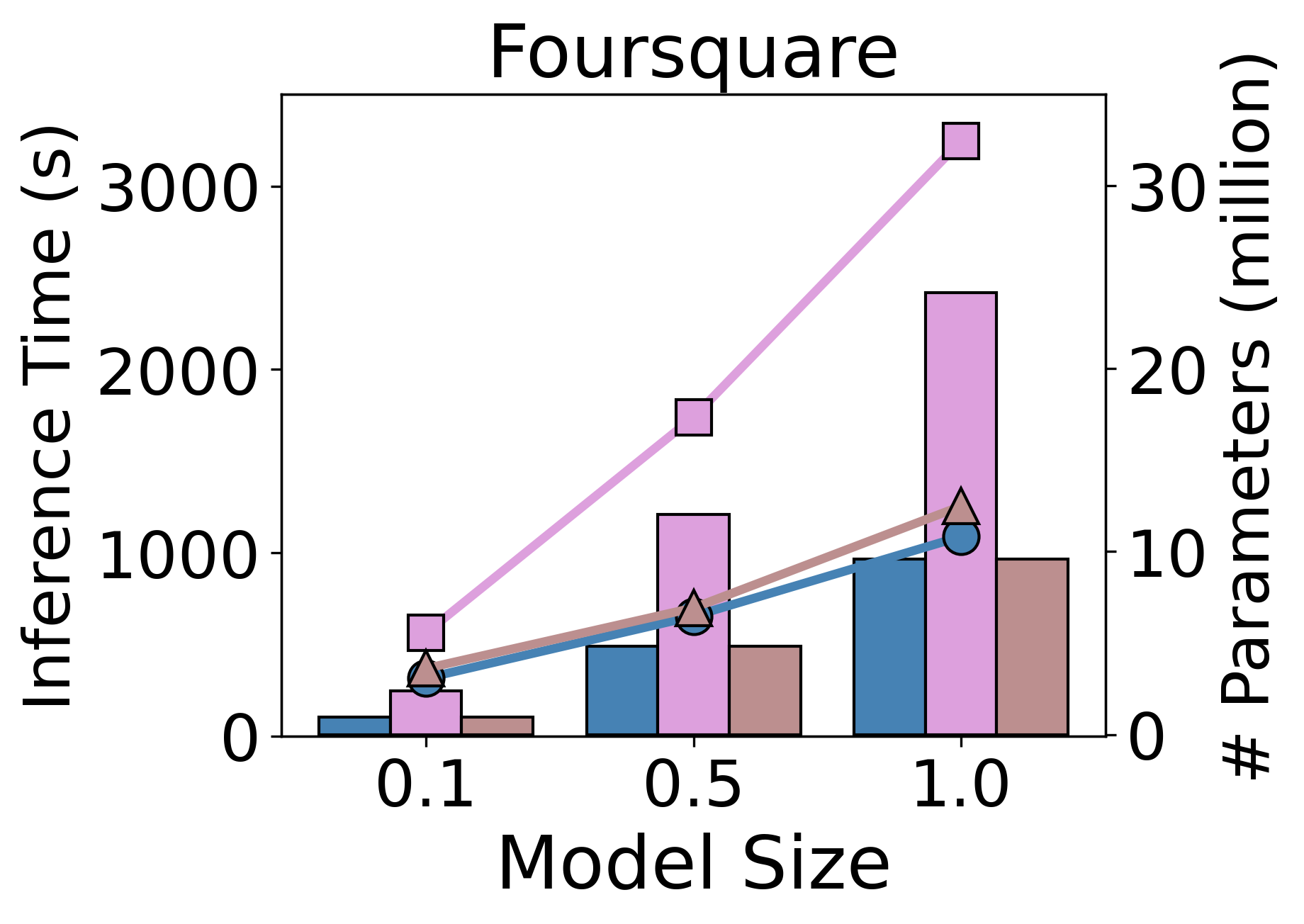}
\end{subfigure}
\caption{Inference time (s) and model size ($\phi$). Inference time denotes the wall time used for generating recommendation list for every user.}
\label{fig:latency}
\end{figure}

For each setting, we increase the model size until the recommendation performance is no longer improved and adopt the model with the best performance as \textbf{Teacher} model.
Then, we build three student models by limiting the model size, i.e.,  $\phi \in \{0.1, 0.5, 1.0\}$.
We call the student model trained without distillation as \textbf{Student} in this section.
Figure \ref{fig:latency} summarizes the model size and inference time.
The inferences are made using PyTorch with CUDA from TITAN Xp GPU and Xeon on Gold 6130 CPU.
It shows that the smaller model has lower inference latency.

\subsubsection{Compared Methods}
We compare the following KD methods distilling the latent knowledge from the intermediate layer of the teacher recommender.
\begin{itemize}[leftmargin=*]
    \item \textbf{FitNet \cite{FitNet}}: A KD method utilizing the original hint regression.
    \item \textbf{Distillation Experts (DE) \cite{DERRD}}: The state-of-the-art KD method distilling the latent knowledge.
    DE elaborates the hint regression.
    \item \textbf{Full Topology Distillation (FTD)}: A proposed method that distills the full topology (Section 3.2).  
    \item \textbf{Hierarchical Topology Distillation (HTD)}: A proposed method that distills the hierarchical topology (Section 3.3).  
\end{itemize}
Note that we do not include the methods distilling the predictions (e.g., RD \cite{RD}, CD \cite{CD}, and RRD \cite{DERRD}) in the competitors, because they are not competing with the methods distilling the latent knowledge \cite{DERRD}.
Instead, we provide experiment results when they are combined with the proposed approach in Section 4.2.

\label{sec:result}

\subsection{Performance Analysis}
\begin{figure}[t]
\centering
\begin{subfigure}[t]{0.7\linewidth}
    \includegraphics[width=\linewidth]{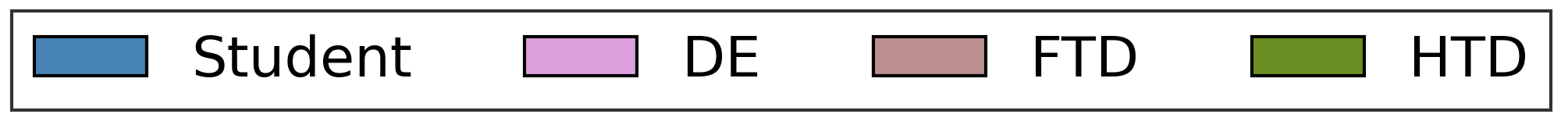}
\end{subfigure}\\
\hspace{-0.345cm}
\begin{subfigure}[t]{0.35\linewidth}
    \includegraphics[width=\linewidth]{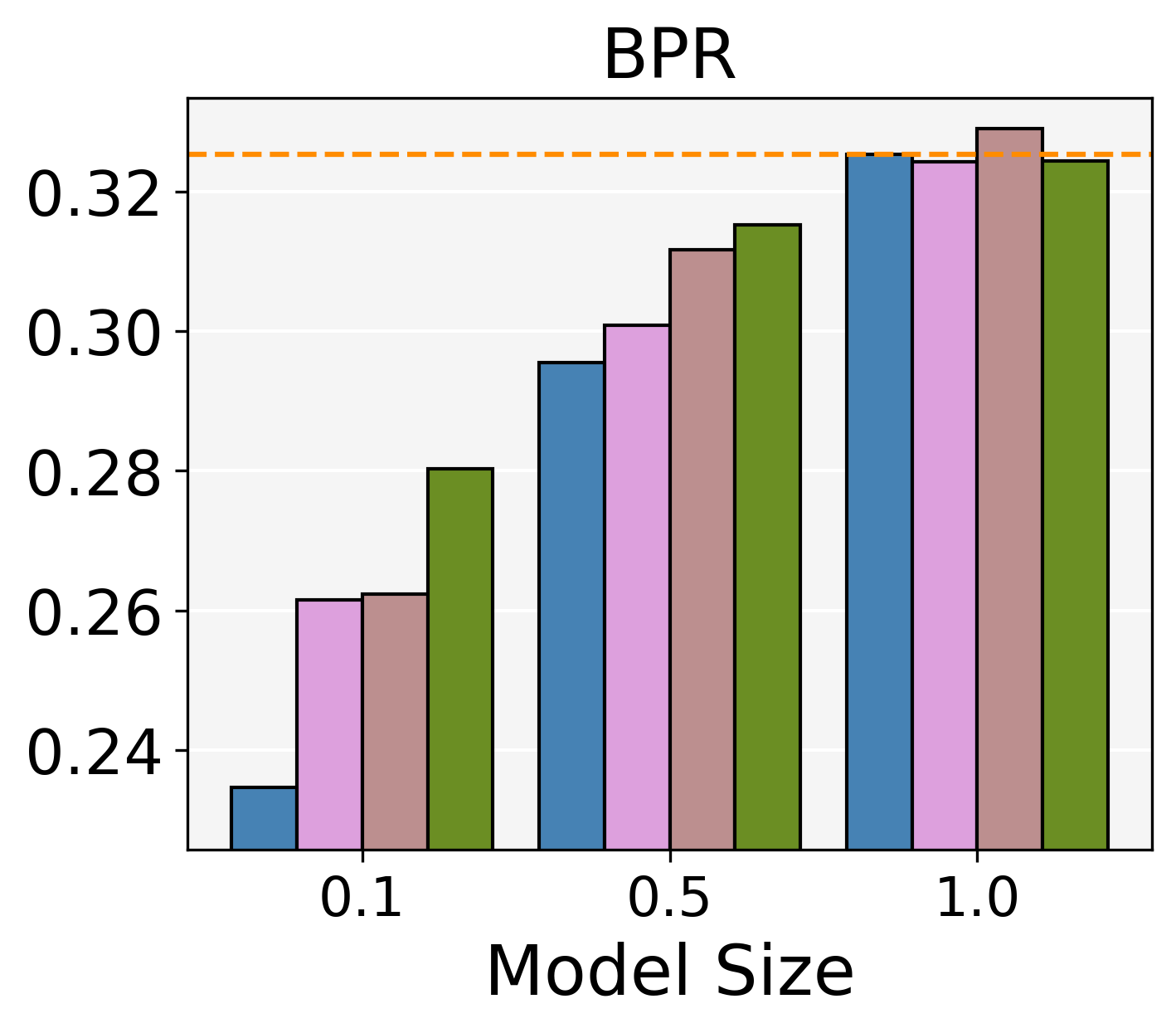}
\end{subfigure}
\hspace{-0.215cm}
\begin{subfigure}[t]{0.36\linewidth}
    \includegraphics[width=\linewidth]{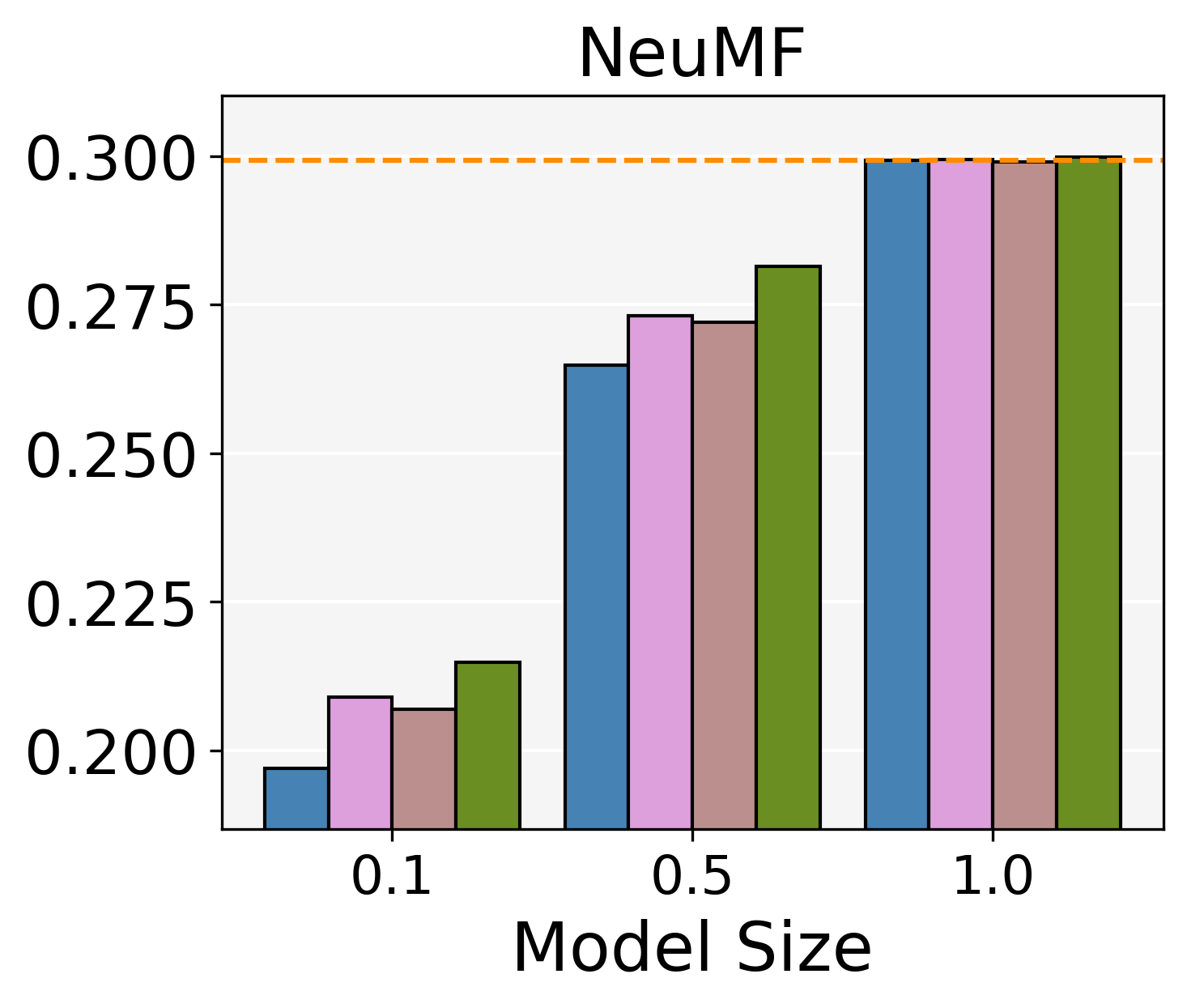}
\end{subfigure}
\hspace{-0.215cm}
\begin{subfigure}[t]{0.35\linewidth}
    \includegraphics[width=\linewidth]{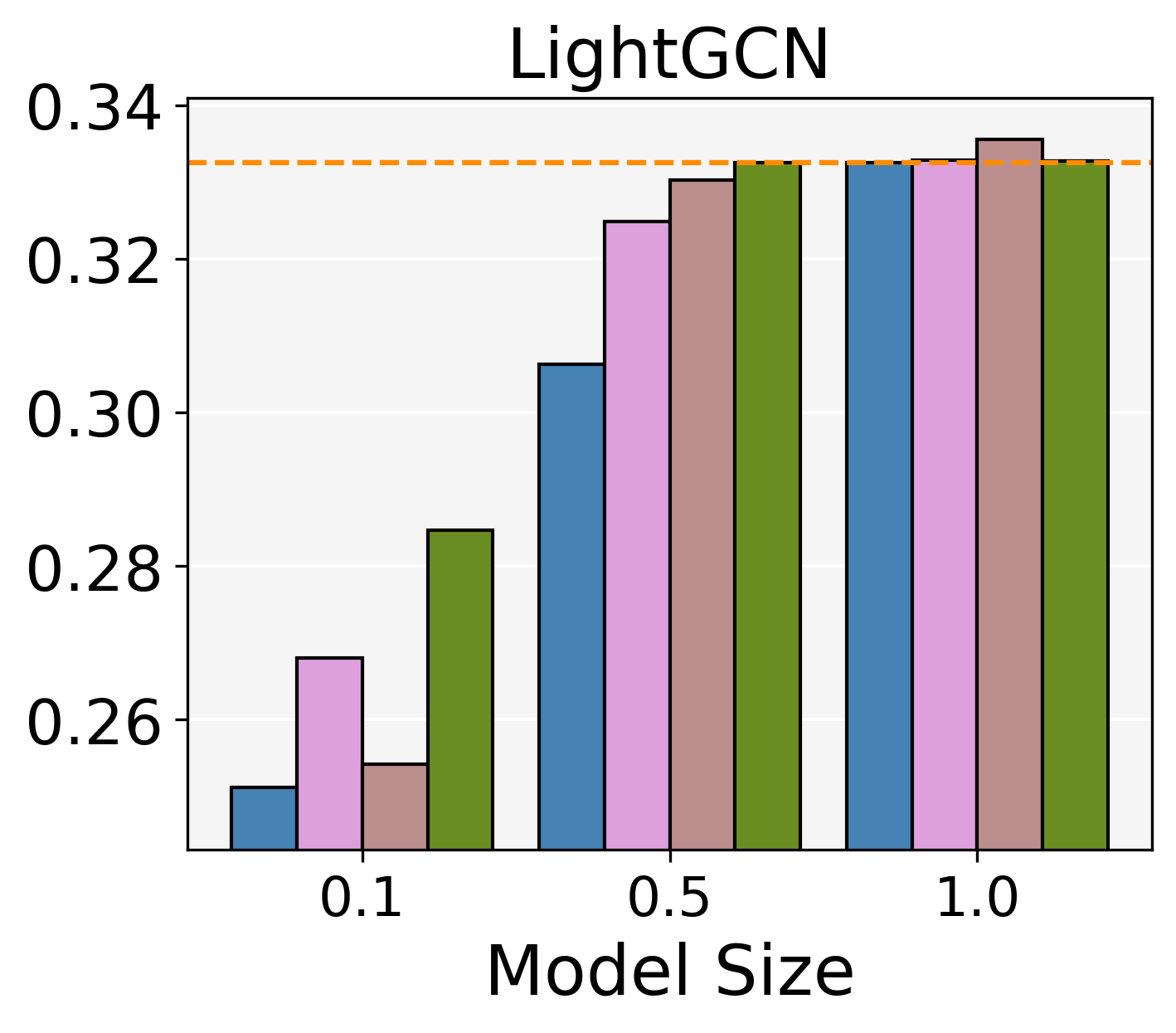}
\end{subfigure}
\hspace{-0.345cm}
\caption*{(a) CiteULike}
\hspace{-0.3cm}
\begin{subfigure}[t]{0.35\linewidth}
    \includegraphics[width=\linewidth]{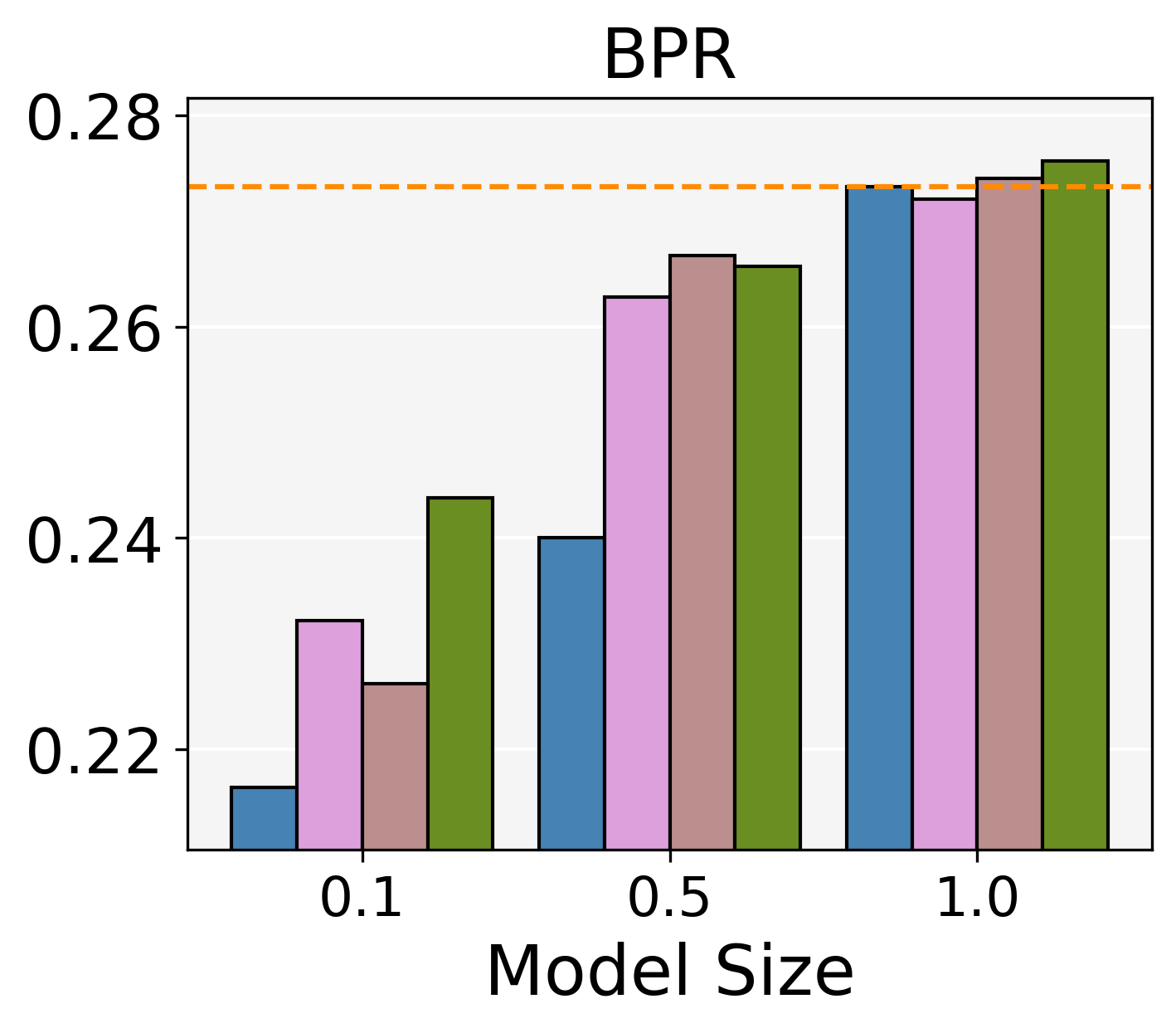}
\end{subfigure}
\hspace{-0.2cm}
\begin{subfigure}[t]{0.35\linewidth}
    \includegraphics[width=\linewidth]{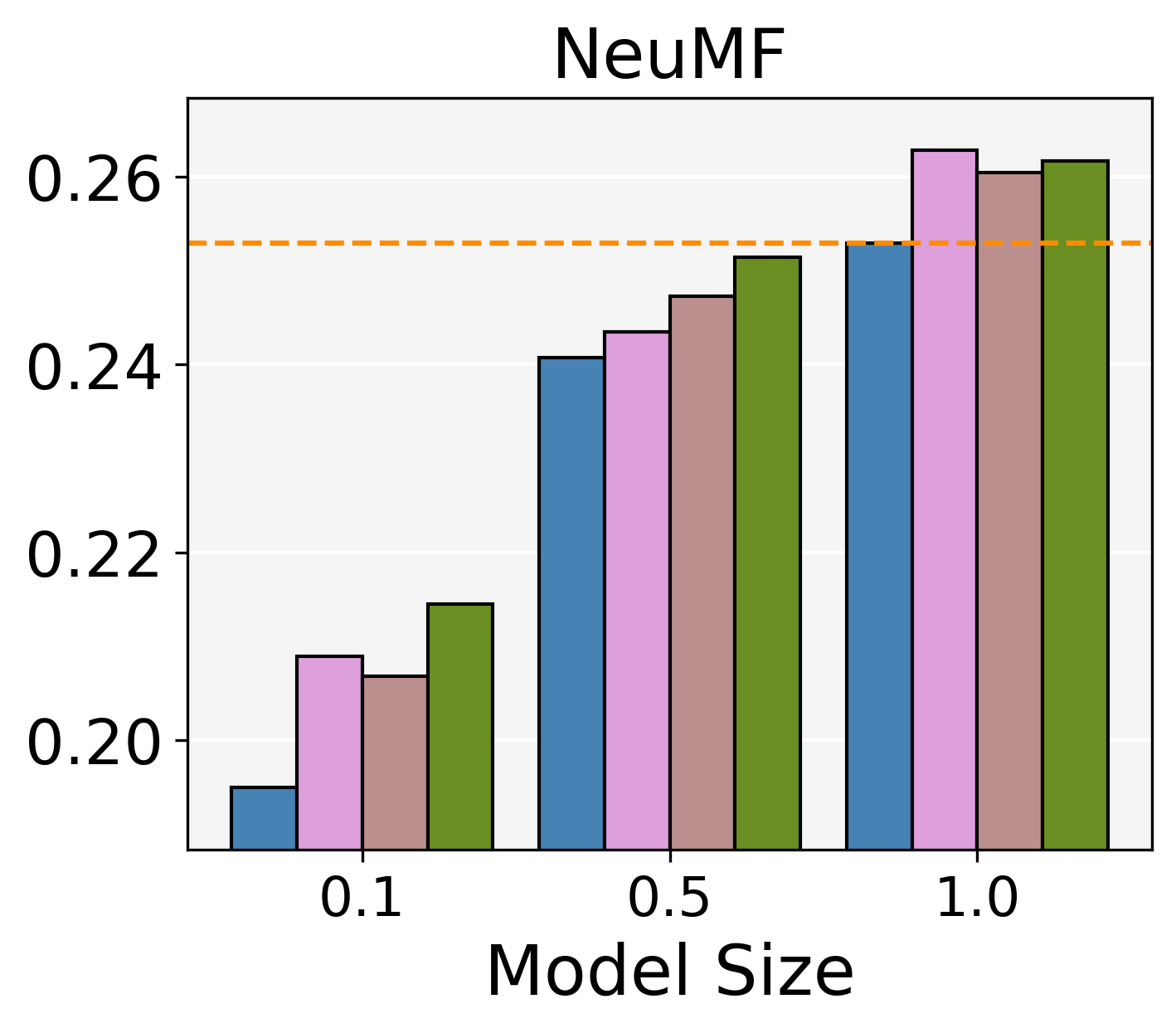}
\end{subfigure}
\hspace{-0.2cm}
\begin{subfigure}[t]{0.35\linewidth}
    \includegraphics[width=\linewidth]{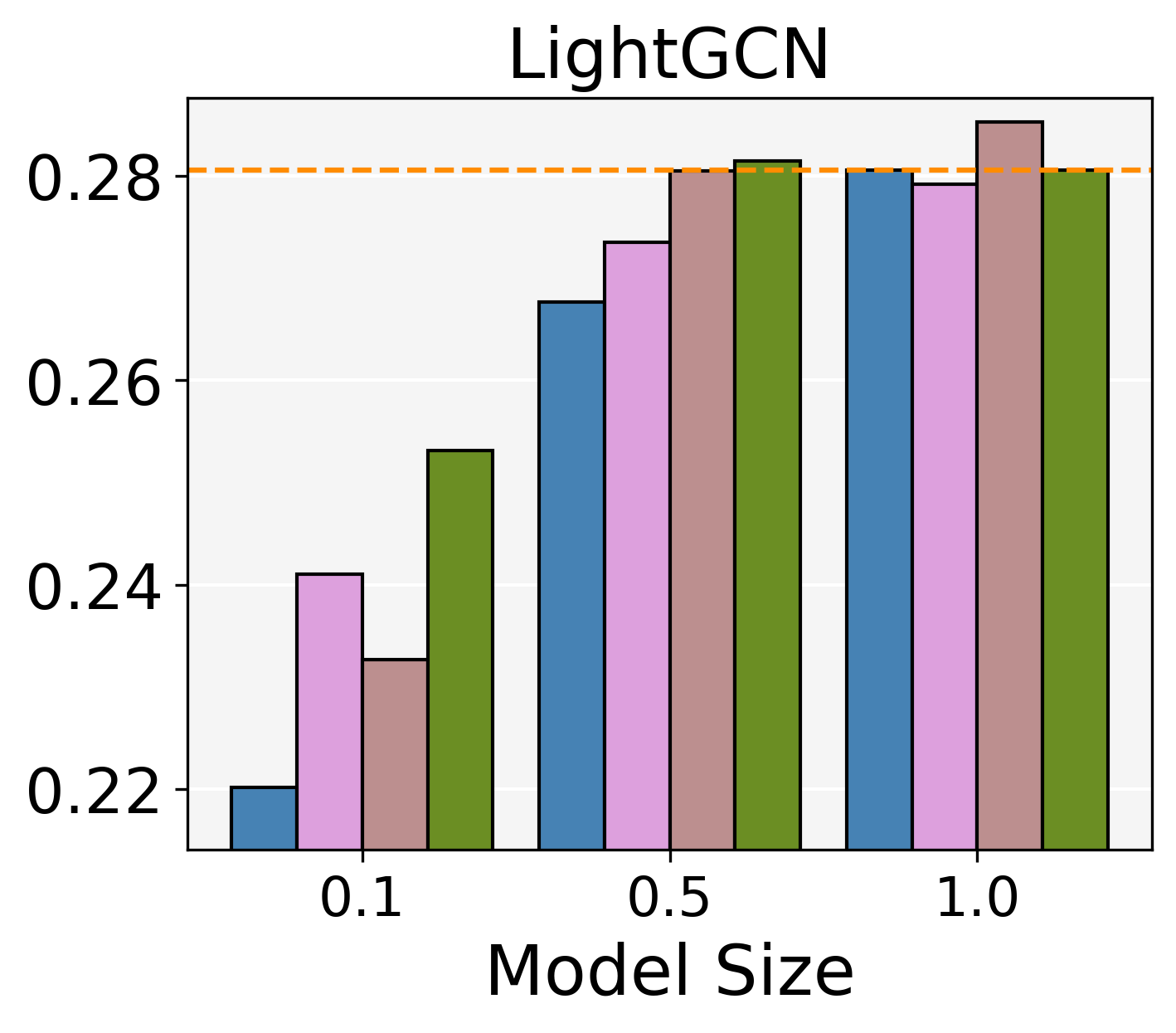}
\end{subfigure}
\hspace{-0.3cm}
\caption*{(b) Foursquare}
\caption{Recall@50 across three different student sizes (Dotted line: Teacher)}
\label{fig:sizes}
\end{figure}

\begin{figure}[t]
\includegraphics[width=0.85\linewidth]{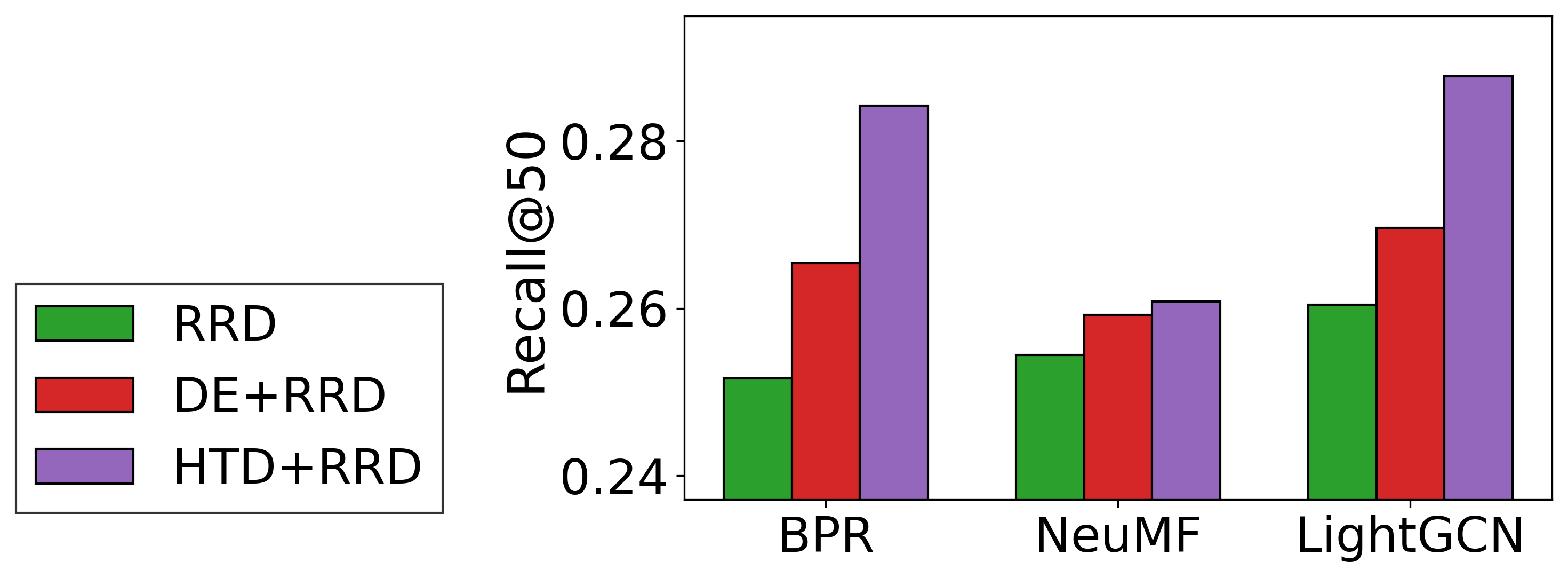}
\caption{Performance comparison with RRD.}
\label{fig:RRD}
\vspace{-0.3cm}
\end{figure}

\begin{figure}[t]
\begin{subfigure}[t]{0.495\linewidth}
    \includegraphics[width=\linewidth]{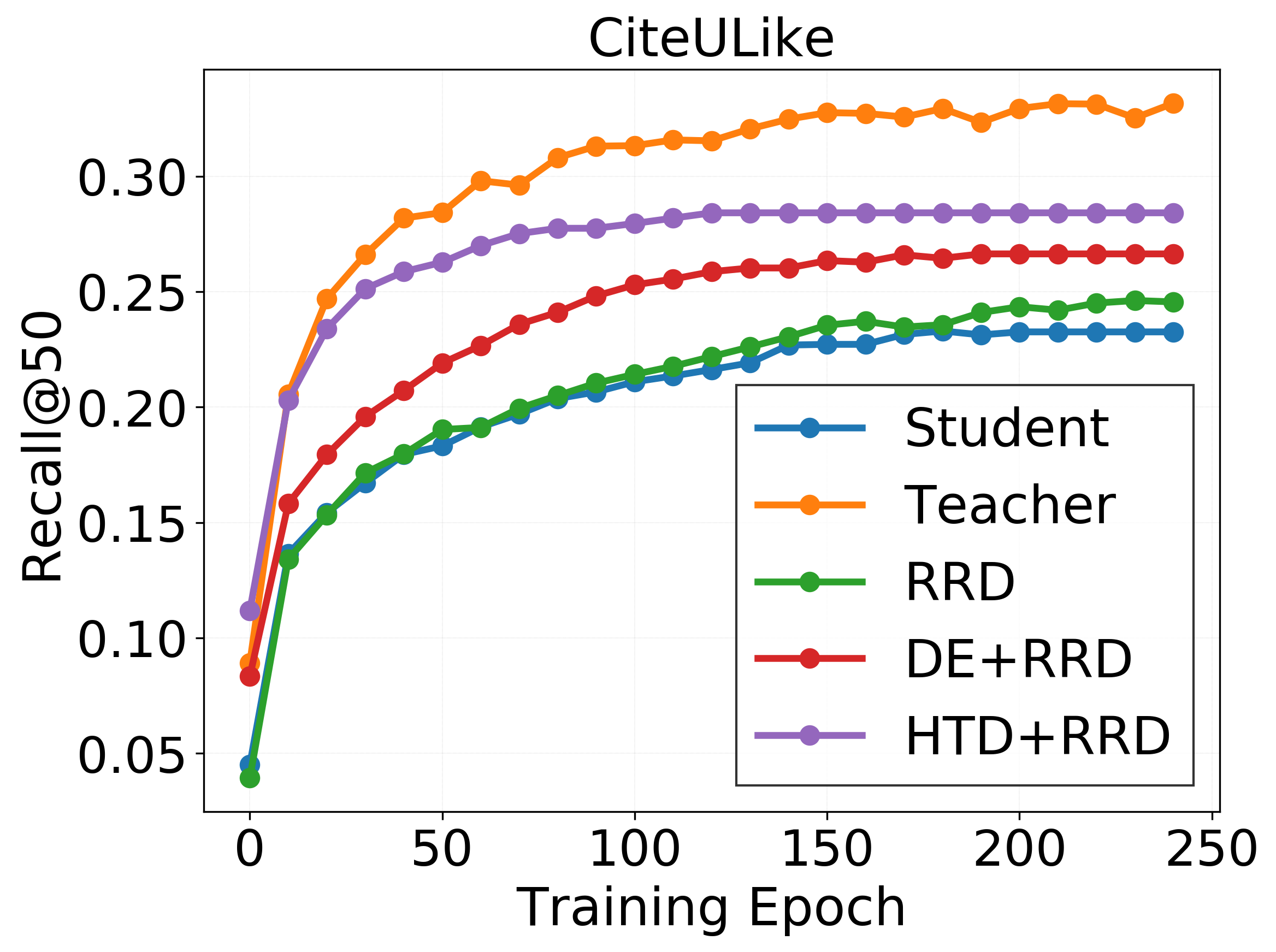}
\end{subfigure}
\begin{subfigure}[t]{0.495\linewidth}
    \includegraphics[width=\linewidth]{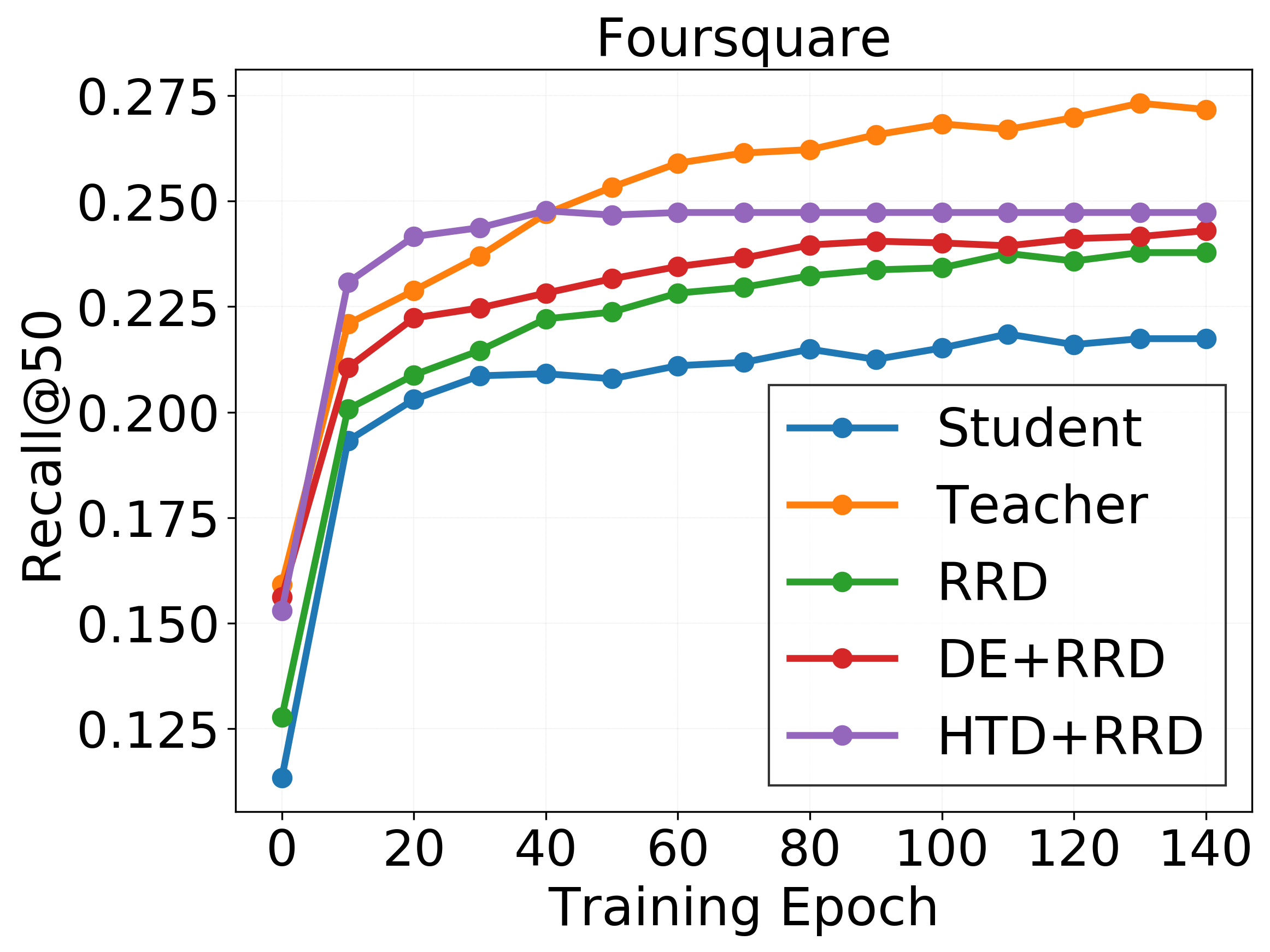}
\end{subfigure}
\caption{Training curves with RRD.}
\label{fig:curve}
\vspace{-0.3cm}
\end{figure}

Table \ref{tbl:maintable} presents top-$N$ recommendation performance of the methods compared ($\phi=0.1$), and Figure \ref{fig:sizes} presents results with three different students sizes. 
For the group-level topology of \proposed, we choose Group(P,e), since it consistently shows better results than Group(P,P).
The detailed comparisons along with other various ablations are reported in Table \ref{tbl:ablation}.
Lastly, results with prediction-based KD method are presented in Figure \ref{fig:RRD} and Figure \ref{fig:curve}.

\newrobustcmd{\B}{\bfseries}
\definecolor{Gray}{gray}{0.95}
\newcolumntype{g}{>{\columncolor{Gray}}c}
\newcolumntype{L}{>{\columncolor{Gray}}l}

\begin{table*}[t]
\centering
\caption{Performance comparison ($\phi=0.1$). \textit{Gain.DE} denotes the improvement of \proposed over DE, \textit{Gain.S} denotes the improvement of \proposed over Student.
\proposed achieves statistically significant improvements over the best baseline. We use the paired t-test with significance level at 0.05 on Recall@50.}
\renewcommand{\arraystretch}{0.25}
\renewcommand{\tabcolsep}{2.5mm}
  \begin{minipage}[t]{1\linewidth}
  \centering
  \begin{tabular}{ccLgggggg}
    \toprule[.15em]
    \rowcolor{white}
    Dataset &Base Model & Method \quad & Recall@10 & NDCG@10 & Recall@20 & NDCG@20 & Recall@50 & NDCG@50 \\
    \midrule[.15em]
    \rowcolor{white}
    & & Teacher & 0.1533 & 0.0883 & 0.2196 & 0.1058 & 0.3253 & 0.1247 \\
    \rowcolor{white}
    &&Student & 0.1014 & 0.0560 & 0.1506 & 0.0684 & 0.2347 & 0.0864 \\
    \rowcolor{white}
    &&FitNet & 0.1097 & 0.0595 & 0.1610 & 0.0738 & 0.2521 & 0.0924 \\
    \rowcolor{white}
    &\multirow{1}{*}{BPR}&DE & 0.1165 & 0.0645 & 0.1696 & 0.0778 & 0.2615 & 0.0960 \\
    \cmidrule{3-9}
    &&FTD & 0.1131 & 0.0630 & 0.1660 & 0.0763 & 0.2624 & 0.0953 \\
    &&HTD & \B 0.1247 & \B 0.0691 & \B 0.1820 & \B 0.0836 & \B 0.2803 & \B 0.1031 \\
    \cmidrule{3-9}
    \rowcolor{white}
    &&\textit{Gain.DE} & 7.0\% & 7.3\% & 7.3\% & 7.5\% & 7.2\% & 7.4\% \\
    \rowcolor{white}
    &&\textit{Gain.S} & 23.0\% & 23.5\% & 20.9\% & 22.3\% & 19.4\% & 19.3\% \\
   \cmidrule[.1em]{2-9}
   \rowcolor{white}
    & & Teacher & 0.1487 & 0.0844 & 0.2048 & 0.0986 & 0.2993 & 0.1155 \\
    \rowcolor{white}
    &&Student & 0.0856 & 0.0449 & 0.1249 & 0.0553 & 0.1970 & 0.0697 \\
    \rowcolor{white}
    &&FitNet & 0.0856 & 0.0469 & 0.1275 & 0.0576 & 0.2020 & 0.0723 \\
    \rowcolor{white}
    \multirow{1}{*}{CiteULike} &\multirow{1}{*}{NeuMF} &DE & 0.0882 & 0.0475 & 0.1306 & 0.0581 & 0.2090 & 0.0736 \\
    \cmidrule{3-9}
    &&FTD & 0.0875 & 0.0474 & 0.1291 & 0.0579 & 0.2069 & 0.0733 \\
    &&HTD & \B 0.0914 & \B 0.0504 & \B 0.1416 & \B 0.0618 & \B 0.2154 & \B 0.0772 \\
    \cmidrule{3-9}
    \rowcolor{white}
    &&\textit{Gain.DE} & 3.6\% & 6.2\% & 8.4\% & 6.4\% & 3.1\% & 4.8\% \\
    \rowcolor{white}
    &&\textit{Gain.S} & 6.8\% & 12.2\% & 13.4\% & 11.8\% & 9.0\% & 10.8\% \\
    \cmidrule[.1em]{2-9}
    \rowcolor{white}
    & & Teacher & 0.1610 & 0.0934 & 0.2274 & 0.1091 & 0.3326 & 0.1299 \\
    \rowcolor{white}
    &&Student & 0.1125 & 0.0618 & 0.1642 & 0.0748 & 0.2512 & 0.0944 \\
    \rowcolor{white}
    &&FitNet & 0.1151 & 0.0642 & 0.1710 & 0.0783 & 0.2653 & 0.0969 \\
    \rowcolor{white}
    &\multirow{1}{*}{LightGCN} &DE & 0.1189 & 0.0664 & 0.1733 & 0.0801 & 0.2680 & 0.0988 \\
    \cmidrule{3-9}
    &&FTD & 0.1112 & 0.0615 & 0.1635 & 0.0747 & 0.2542 & 0.0926 \\
    &&HTD & \B 0.1322 & \B 0.0742 & \B 0.1902 & \B 0.0888 & \B 0.2847 & \B 0.1075 \\
    \cmidrule{3-9}
    \rowcolor{white}
   & &\textit{Gain.DE} & 11.2\% & 11.8\% & 9.7\% & 10.9\% & 6.2\% & 8.8\% \\
   \rowcolor{white}
    &&\textit{Gain.S} & 17.5\% & 20.1\% & 15.8\% & 18.7\% & 13.3\% & 13.9\% \\    
    \midrule[.15em]
    \rowcolor{white}
    & & Teacher & 0.1187 & 0.0695 & 0.1700 & 0.0825 & 0.2732 & 0.1028 \\
    \rowcolor{white}
    &&Student & 0.0911 & 0.0544 & 0.1333 & 0.0648 & 0.2164 & 0.0809 \\
    \rowcolor{white}
    &&FitNet & 0.0957 & 0.0564 & 0.1386 & 0.0672 & 0.2258 & 0.0845 \\
    \rowcolor{white}
    &\multirow{1}{*}{BPR} &DE & 0.0979 & 0.0567 & 0.1434 & 0.0681 & 0.2322 & 0.0856 \\
   \cmidrule{3-9}
    &&FTD & 0.0987 & 0.0582 & 0.1417 & 0.0690 & 0.2262 & 0.0857 \\
    &&HTD & \B 0.1037 & \B 0.0622 & \B 0.1505 & \B 0.0740 & \B 0.2438 & \B 0.0921 \\
    \cmidrule{3-9}
    \rowcolor{white}
    &&\textit{Gain.DE} & 5.9\% & 9.7\% & 5.0\% & 8.7\% & 5.0\% & 7.6\% \\
    \rowcolor{white}
   & &\textit{Gain.S} & 13.8\% & 14.3\% & 12.9\% & 14.2\% & 12.7\% & 13.8\% \\
    \cmidrule[.1em]{2-9}
    \rowcolor{white}
    & & Teacher & 0.1060 & 0.0590 & 0.1546 & 0.0716 & 0.2529 & 0.0910 \\
    \rowcolor{white}
    &&Student & 0.0737 & 0.0393 & 0.1125 & 0.0490 & 0.1950 & 0.0653 \\
    \rowcolor{white}
    &&FitNet & 0.0829 & 0.0462 & 0.1243 & 0.0564 & 0.2062 & 0.0729 \\
    \rowcolor{white}
    \multirow{1}{*}{Foursquare} &\multirow{1}{*}{NeuMF} &DE & 0.0855 & 0.0476 & 0.1255 & 0.0576 & 0.2089 & 0.0741 \\
    \cmidrule{3-9}
    &&FTD & 0.0823 & 0.0451 & 0.1233 & 0.0554 & 0.2068 & 0.0719 \\
    &&HTD & \B 0.0891 & \B 0.0501 & \B 0.1294 & \B 0.0601 & \B 0.2152 & \B 0.0770 \\
    \cmidrule{3-9}
    \rowcolor{white}
   & &\textit{Gain.DE} & 4.3\% & 5.3\% & 3.1\% & 4.3\% & 3.0\% & 3.9\% \\
   \rowcolor{white}
   & &\textit{Gain.S} & 20.9\% & 27.2\% & 15.0\% & 22.7\% & 10.0\% & 17.9\% \\
    \cmidrule[.1em]{2-9}
    \rowcolor{white}
   &  & Teacher & 0.1259 & 0.0730 & 0.1779 & 0.0865 & 0.2806 & 0.1067 \\
   \rowcolor{white}
   & &Student & 0.0951 & 0.0564 & 0.1372 & 0.0670 & 0.2202 & 0.0834 \\
   \rowcolor{white}
   & &FitNet & 0.0993 & 0.0587 & 0.1431 & 0.0697 & 0.2315 & 0.0872 \\
   \rowcolor{white}
   & \multirow{1}{*}{LightGCN} &DE & 0.1051 & 0.0617 & 0.1503 & 0.0731 & 0.2410 & 0.0910 \\
    \cmidrule{3-9}
   & &FTD & 0.1018 & 0.0602 & 0.1466 & 0.0714 & 0.2327 & 0.0884 \\
   & &HTD & \B 0.1119 & \B 0.0652 & \B 0.1597 & \B 0.0772 & \B 0.2531 & \B 0.0956 \\
    \cmidrule{3-9}
    \rowcolor{white}
   & &\textit{Gain.DE} & 6.4\% & 5.6\% & 6.3\% & 5.6\% & 5.0\% & 5.1\% \\
   \rowcolor{white}
   & &\textit{Gain.S} & 17.7\% & 15.6\% & 16.4\% & 15.2\% & 14.9\% & 14.6\% \\
    \bottomrule[.15em]
  \end{tabular}
  \end{minipage}
    \label{tbl:maintable}
    \vspace{-0.2cm}
\end{table*}

\begin{table}[t!]
\small
\centering
\caption{Performance comparison with ablations ($\phi=0.1$).}
\renewcommand{\arraystretch}{0.72}
\renewcommand{\tabcolsep}{0.7mm}
  \begin{minipage}[t]{1\linewidth}
  \centering
  \begin{tabular}{clcc cc}
    \toprule[.15em]
     \multirow{2}{*}{\makecell{Base\\Model}} & \multirow{2}{*}{Method}  &  \multicolumn{2}{c}{CiteULike}& \multicolumn{2}{c}{Foursquare}\\
    \cmidrule(lr){3-4}\cmidrule(lr){5-6}
     &  &Recall@50 & NDCG@50 & Recall@50 & NDCG@50 \\
    \midrule[.15em]
     &FTD &  0.2624 &  0.0953 &  0.2262 & 0.0857\\
    &FTD+DE  & 0.2650 & 0.0969  & 0.2339 & 0.0867 \\
    \cmidrule{2-6}
    \multirow{1}{*}{BPR}& HTD  & \B 0.2803 & \B 0.1031  & \B 0.2438 & \B 0.0921 \\
    &Group only  & 0.2619 & 0.0948  & 0.2349 & 0.0874 \\
    &Entity only  & 0.2608 & 0.0968  & 0.2361 & 0.0871 \\
    &Group (P,P) & 0.2648 & 0.0982  & 0.2399 & 0.0887 \\
    \cmidrule{1-6}
     &FTD & 0.2542 &  0.0926 &  0.2327 & 0.0884\\
    &FTD+DE & 0.2572 & 0.0931  & 0.2335 & 0.0872 \\
    \cmidrule{2-6}
    \multirow{1}{*}{LightGCN}& HTD  & \B 0.2847 & \B 0.1075 & \B 0.2531 & \B 0.0956 \\
    &Group only & 0.2596  & 0.0976   & 0.2432  & 0.0910 \\
    &Entity only  & 0.2709  & 0.1010   & 0.2453  & 0.0910 \\
    &Group (P,P)  & 0.2683  & 0.0987    & 0.2459  & 0.0909 \\
    \bottomrule[.15em]
  \end{tabular}
  \end{minipage}
    \label{tbl:ablation}
    \vspace{-0.3cm}
\end{table}

\vspace{0.1cm}
\noindent
\textbf{Overall Evaluation.} 
In Table \ref{tbl:maintable}, we observe that \proposed achieves significant performance gains compared to the main competitor, i.e., DE.
This result shows that distilling the relational knowledge provides better guidance than only distilling the knowledge of individual representation.
We also observe that FTD is not always effective and sometimes even degrades the student’s recommendation performance (e.g., LightGCN on CiteULike).
As the student's size is highly limited compared to the teacher in the KD scenario, learning all the relational knowledge is daunting for the student, which leads to degrade the effects of distillation.
This result supports our claim that the relational knowledge should be distilled considering the huge capacity gap.
Also, the results show that \proposed successfully copes with the issue, enabling the student to effectively learn the relational knowledge.

In Figure \ref{fig:sizes}, we observe that as the model size increases, the performance gap between FTD and \proposed decreases. 
FTD achieves comparable and even higher performance than \proposed when the student has enough capacity (e.g., LightGCN on Foursquare with $\phi=1.0$).
This result again verifies that the effectiveness of the proposed topology distillation approach.
It also shows that FTD can be applied to maximize the performance of recommender in the scenario where there is no constraint on the model size by self-distillation.

\vspace{0.1cm}
\noindent
\textbf{Comparison with ablations.} 
In Table \ref{tbl:ablation}, we provide the comparison with diverse ablations.
For FTD, we report the results when it is used with the state-of-the-art hint regression method (denoted as \textbf{FTD+DE}).
As HTD includes DE for the group assignment, comparison with FTD+DE shows the direct impacts of topology relaxation.
We observe that FTD+DE is not as effective as \proposed and sometimes even achieves worse performance than~DE.
This shows that the power of \proposed comes from the distillation strategy transferring the relational knowledge in multi-levels.

For HTD, we compare three ablations: \textbf{1) Group only} that considers only group-level topology, \textbf{2) Entity only} that considers only entity-level topology, and \textbf{3) Group (P,P)} that considers the relations of the prototypes only for the group-level topology (Section 3.3.2).
We first observe that both group-level and entity-level topology are indeed necessary.
Without either of them, the performance considerably drops.
The group-level topology includes the summarized relations across the groups, providing the overview of the entire topology.
On the other hand, the entity-level topology includes the full relations in each group, providing fine-grained supervision of how the entities should be correlated.
Based on both two-level topology, \proposed effectively transfers the relational knowledge.
Lastly, we observe that \text{Group (P,P)} is not as effective as \text{Group(P,e)} adopted in \proposed.
We conjecture that summarizing numerous relations across the groups into a single relation may lose too much information and cannot effectively boost the student.

\vspace{0.1cm}
\noindent
\textbf{With prediction-based KD method.} 
We report the results with the state-of-the-art prediction KD method (i.e., RRD \cite{DERRD}) on CiteULike with $\phi=0.1$ in Figure \ref{fig:RRD}.
Also, we provide the training curves of BPR with $\phi=0.1$ in Figure \ref{fig:curve}\footnote{After the early stopping on the validation set, we plot the final performance.}.
First, we observe that the effectiveness of RRD is considerably improved when it is applied with the KD method distilling the latent knowledge (i.e., DE and HTD).
This result aligns with the results reported in \cite{DERRD} and shows the importance of distilling the latent knowledge.
Second, we observe that the student recommender achieves the best performance with \proposed.
Unlike RRD, which makes the student imitate the ranking order of items in each user's recommendation list, \proposed distills relations existing in the representation space via the topology matching.
The topology includes much rich supervision including not only user-item relations but also \textit{user-user} relations and \textit{item-item} relations.
By obtaining this additional information in a proper manner considering the capacity gap, the student can be fully improved with \proposed.

\subsection{Benefit of Topology Distillation}
To further ascertain the benefit of the topology distillation, we provide in-depth analysis on representations obtained by each KD~method ($\phi=0.1$).

\begin{figure}[t]
\begin{subfigure}[t]{0.6\linewidth}
    \includegraphics[width=\linewidth]{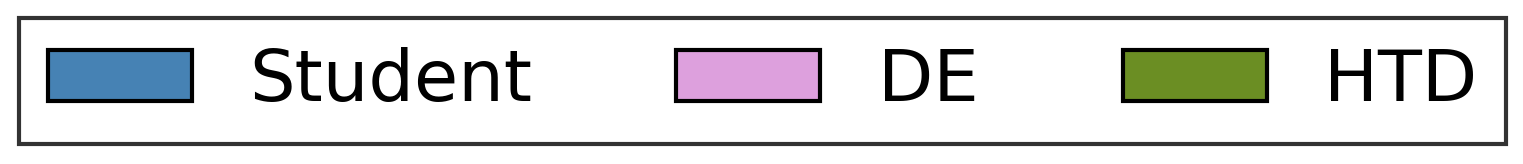}
\end{subfigure}\\
\hspace{-0.2cm}
\begin{subfigure}[t]{0.5\linewidth}
    \includegraphics[width=\linewidth]{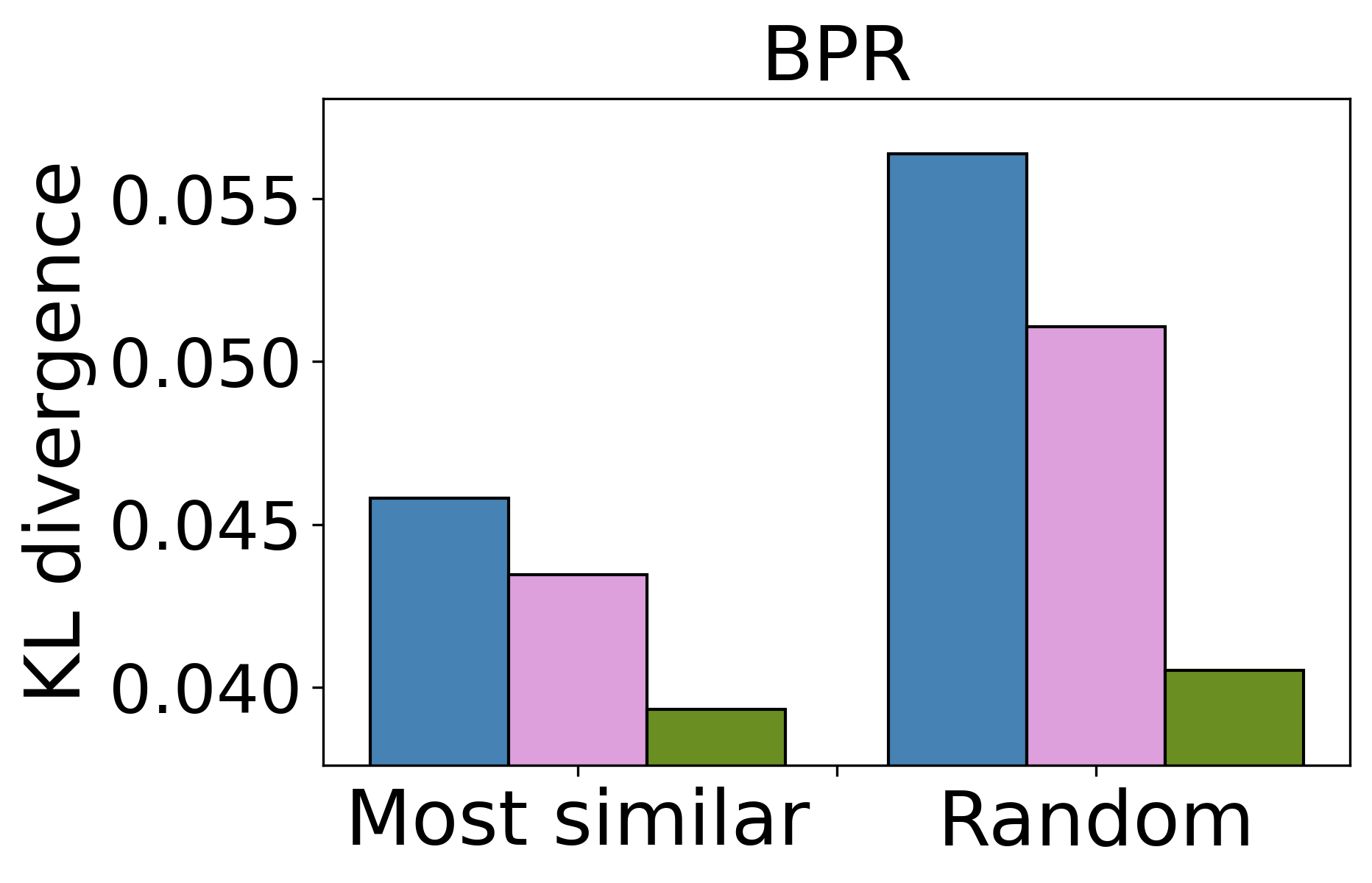}
\end{subfigure}
\hspace{-0.2cm}
\begin{subfigure}[t]{0.5\linewidth}
    \includegraphics[width=\linewidth]{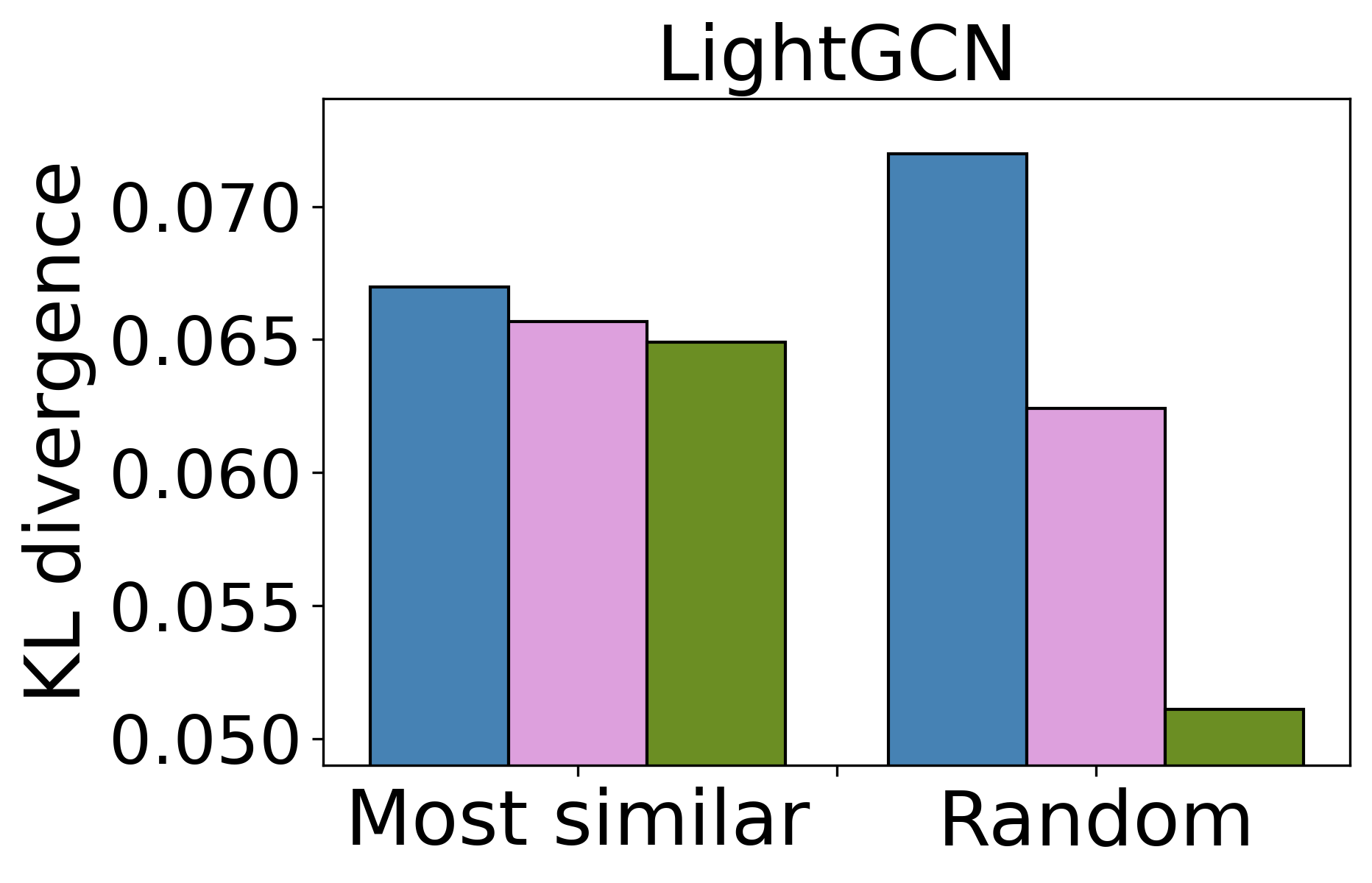}
\end{subfigure}
\hspace{-0.2cm}
\caption{The average KL divergence from similarity distributions obtained in the teacher representation~spaces.}
\label{fig:kld}
\vspace{-0.3cm}
\end{figure}

First, we evaluate whether the topology distillation indeed makes the student better preserve the relations in the teacher representation space than the existing method.
For quantitative evaluation, we conduct the following steps:
\textbf{1)} In the teacher space, for each representation, we compute the similarity distributions with 100 \textit{most similar} representations and 100 \textit{randomly selected} representations, respectively.
\textbf{2)} In the student space, for each representation, we compute the similarity distributions with the representations chosen in the teacher space.
\textbf{3)} We compute KL divergence for each distribution and report the average value in Figure \ref{fig:kld}.
KL divergence of `Most similar' indicates how well the detailed relations among the strongly correlated representations are preserved, and that of `Random' indicates how well the overall relations in the space are preserved.
We observe that \proposed achieves the lowest KL divergence for both Most similar and Random, which shows that \proposed indeed enables the student to better preserve the relations in the teacher~space.

Second, we compare the performance of two downstream tasks that evaluate how well each method encodes the items’ characteristics (or semantics) into the representations.
We perform the tag retrieval task for CiteULike and the region classification task for Foursquare.
We train a linear and a non-linear model to predict the tag/region of each item by using the fixed item representation as the input.
The detailed setup is provided in the appendix. 
In Table \ref{tbl:downstream_tasks}, we observe that \proposed achieves consistently higher performance than DE on both of two downstream tasks.
This strongly indicates that the representation space induced by \proposed more accurately captures the item’s semantics compared to the space induced by~DE.

In sum, with the topology distillation approach, the student can indeed better preserve the relations in the teacher space.
This not only improves the recommendation performance but also allows it to better capture the semantic of entities.

\begin{table}[t]
\centering
\small
\caption{Performance comparison on downstream tasks.}
\renewcommand{\arraystretch}{0.7}
\renewcommand{\tabcolsep}{1.0mm}
  \centering
  \begin{tabular}{cclcc}
    \toprule[.15em]
    \makecell{Base\\Model} & Type & Method & \makecell{Tag Retrieval \\(Recall@10)}& \makecell{Region Classification \\(Accuracy)} \\
    \midrule[.15em]
    \multirow{8}{*}{\small BPR}&\multirow{4}{*}{linear} & Teacher & 0.3121$\pm$2.7e-3& 0.6531$\pm$2.5e-3 \\
    &&Student & 0.2421$\pm$1.7e-3& 0.4222$\pm$7.4e-3\\
    &&DE & 0.2542$\pm$1.5e-3& 0.5448$\pm$3.2e-3 \\
    &&HTD & \textbf{0.2635}$\pm$1.2e-3& \textbf{0.5653}$\pm$5.0e-3  \\
    \cmidrule{2-5}
    &\multirow{4}{*}{\makecell{non-linear}} & Teacher & 0.3123$\pm$1.4e-3& 0.6357$\pm$8.8e-3  \\
    &&Student & 0.2701$\pm$2.1e-3&0.4224$\pm$7.9e-3  \\
    &&DE & 0.2807$\pm$2.5e-3& 0.5123$\pm$6.6e-3\\
    &&HTD & \textbf{0.2909}$\pm$2.2e-3&\textbf{0.5318}$\pm$3.5e-3  \\
    \midrule
    \multirow{8}{*}{\small LightGCN}&\multirow{4}{*}{linear} & Teacher & 0.3489$\pm$0.6e-3& 0.6787$\pm$6.1e-3 \\
    &&Student & 0.2523$\pm$0.7e-3& 0.4635$\pm$3.1e-3\\
    &&DE & 0.2565$\pm$1.2e-3& 0.5654$\pm$6.5e-3 \\
    &&HTD & \textbf{0.2650}$\pm$0.3e-3&\textbf{0.5854}$\pm$7.7e-3  \\
    \cmidrule{2-5}
    &\multirow{4}{*}{non-linear} & Teacher & 0.3360$\pm$1.2e-3& 0.6504$\pm$5.9e-3\\
    &&Student & 0.2971$\pm$0.3e-3& 0.4354$\pm$6.5e-3 \\
    &&DE & 0.3053$\pm$1.3e-3& 0.5250$\pm$3.4e-3 \\
    &&HTD & \textbf{0.3138}$\pm$1.1e-3& \textbf{0.5485}$\pm$8.1e-3\\
    \bottomrule[.15em]
  \end{tabular}
    \label{tbl:downstream_tasks}
    \vspace{-0.3cm}
\end{table}
\begin{figure}[t]
\hspace{-0.3cm}
\begin{subfigure}[t]{0.262\linewidth}
    \includegraphics[width=\linewidth]{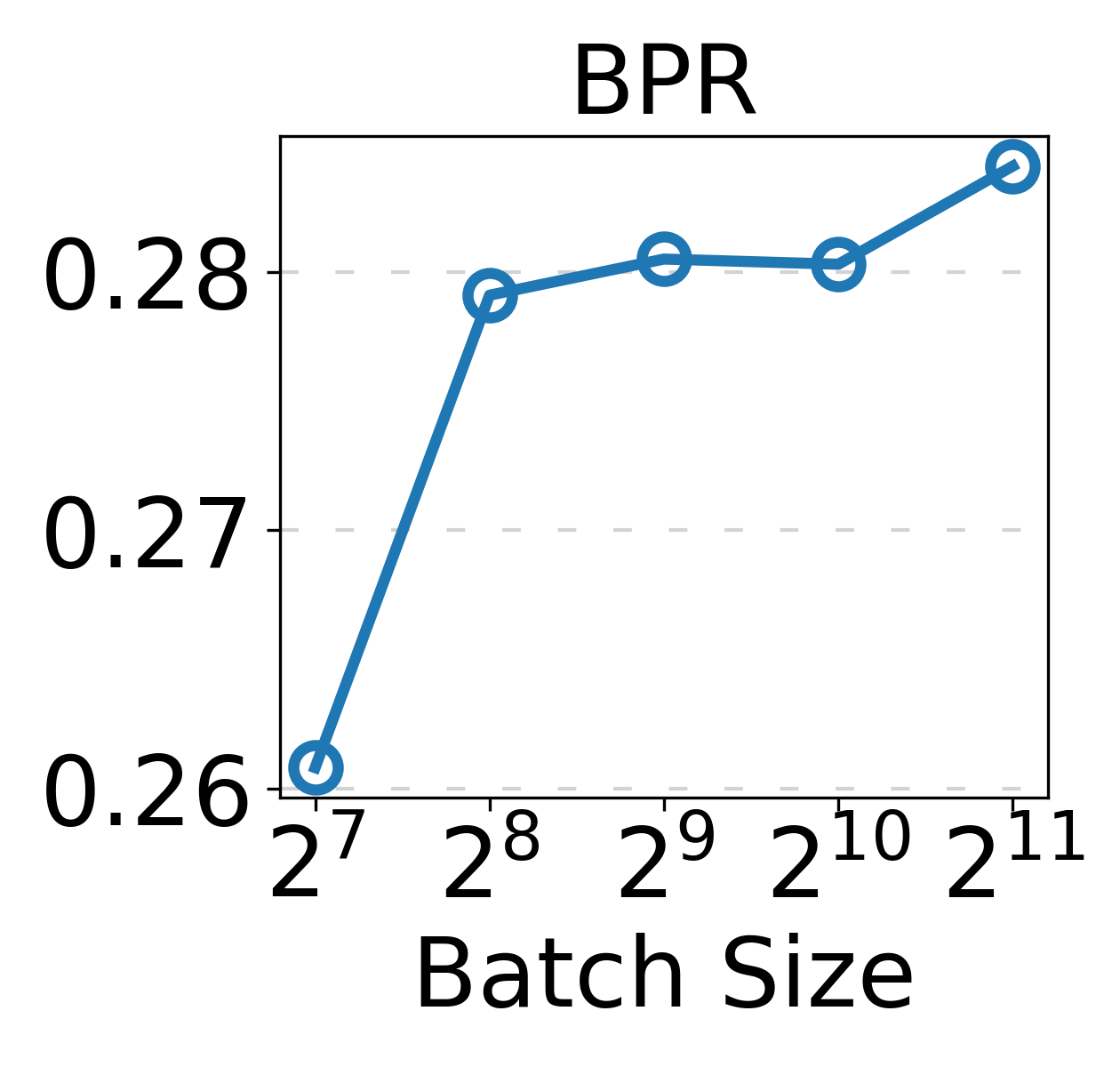}
\end{subfigure}
\hspace{-0.22cm}
\begin{subfigure}[t]{0.262\linewidth}
    \includegraphics[width=\linewidth]{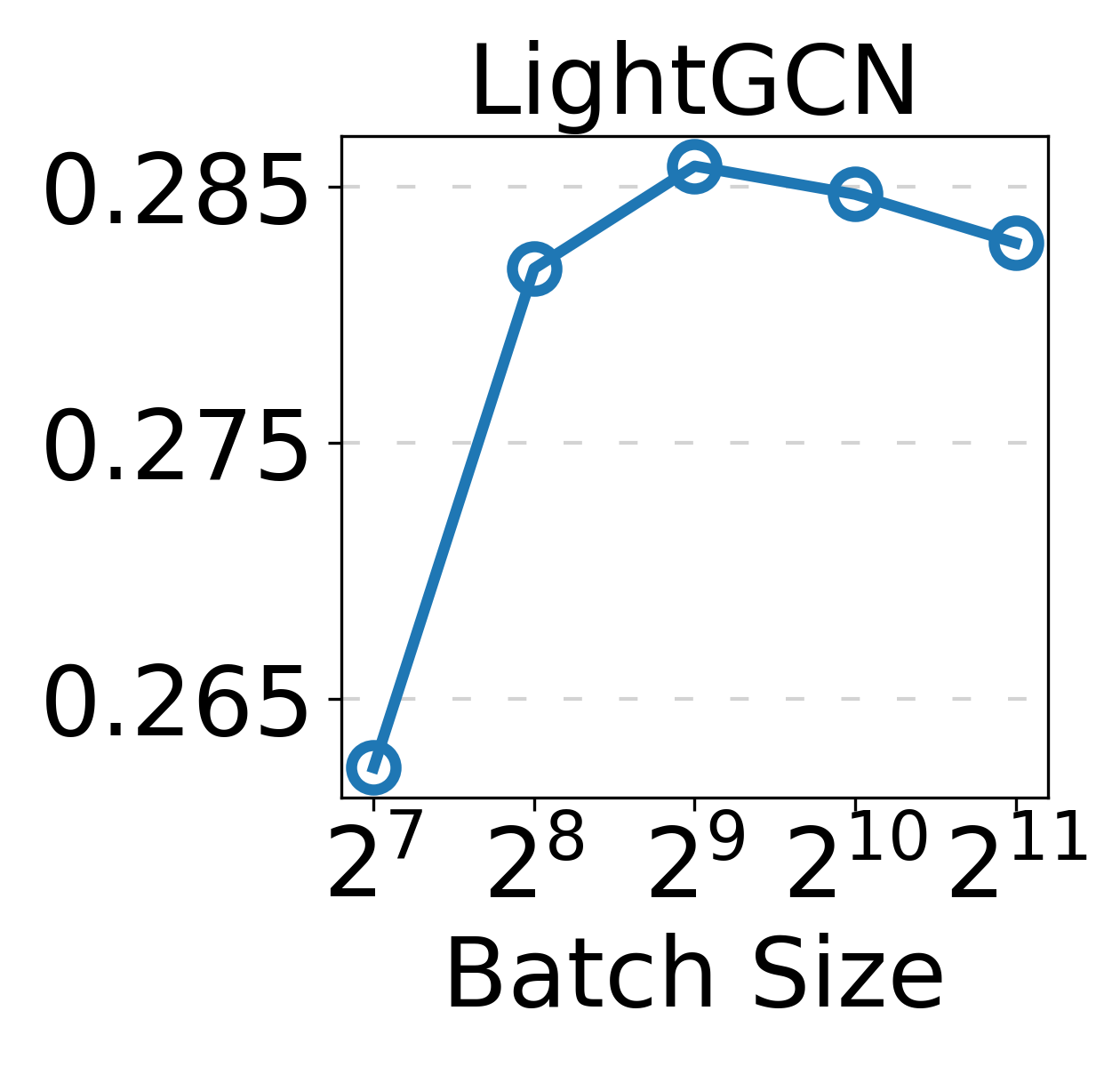}
\end{subfigure}
\hspace{-0.22cm}
\begin{subfigure}[t]{0.262\linewidth}
    \includegraphics[width=\linewidth]{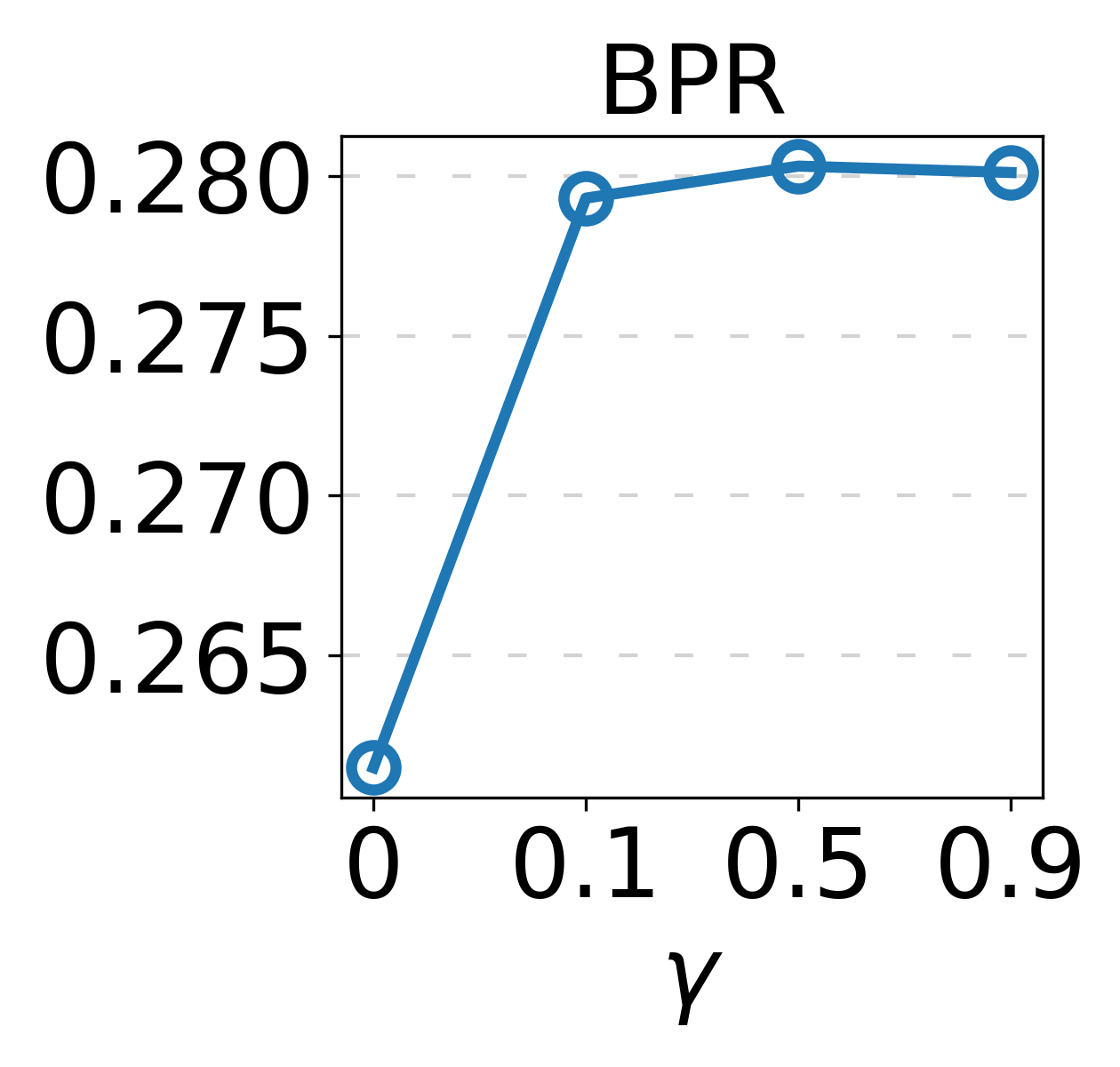}
\end{subfigure}
\hspace{-0.22cm}
\begin{subfigure}[t]{0.262\linewidth}
    \includegraphics[width=\linewidth]{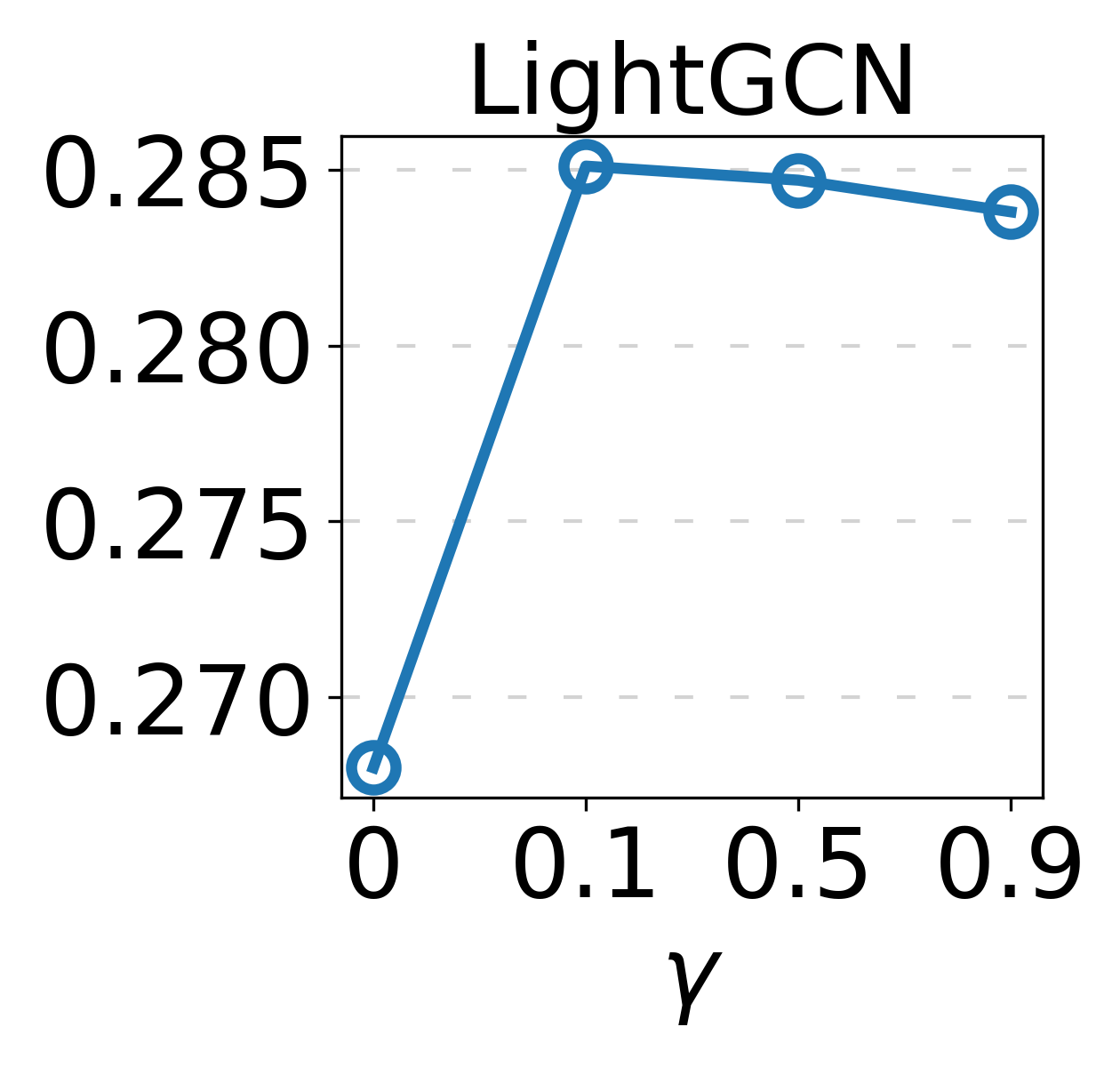}
\end{subfigure}
\hspace{-0.3cm}
\caption*{(a) Effects of batch size and $\gamma$}
\begin{subfigure}[t]{0.4\linewidth}
    \includegraphics[width=\linewidth]{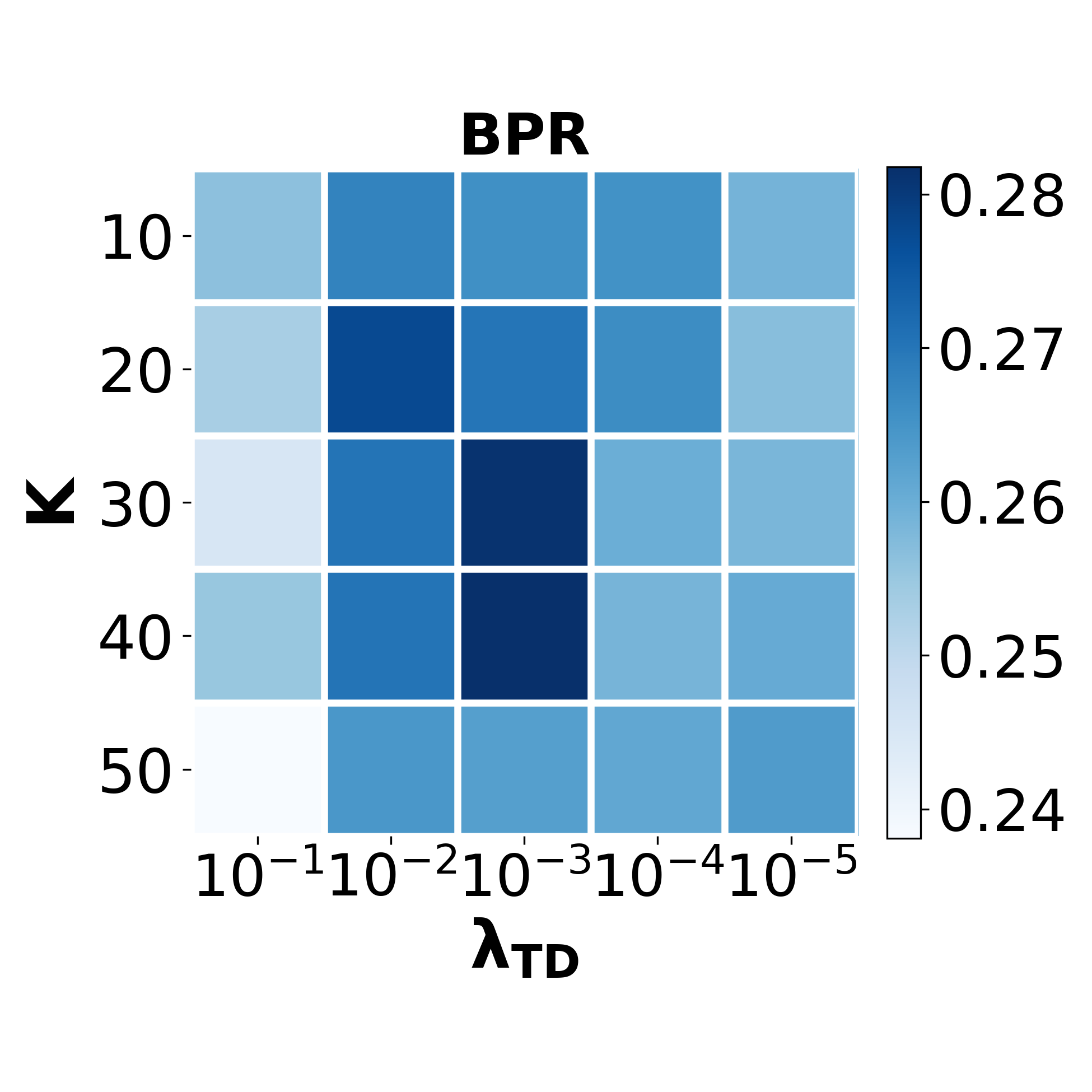}
\end{subfigure}
\hspace{0.35cm}
\begin{subfigure}[t]{0.4\linewidth}
    \includegraphics[width=\linewidth]{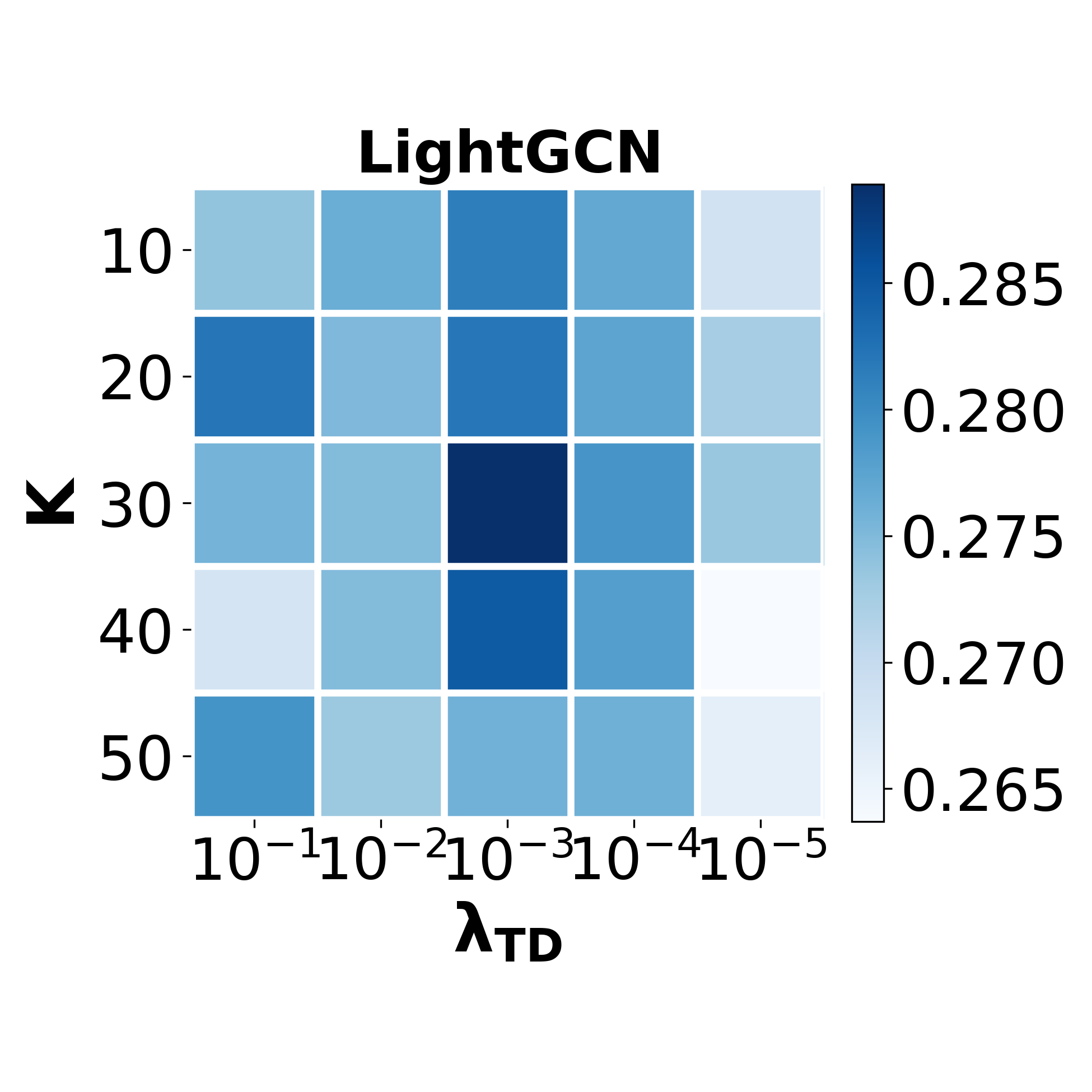}
\end{subfigure}
\vspace{-0.1cm}
\caption*{(b) Effects of $\lambda_{TD}$ and $K$}
\caption{Effects of the hyperparameters (Recall@50).}
\label{fig:hp}
\end{figure}

\subsection{Hyperparameter Analysis}
We provide analyses to guide the hyperparameter selection of the topology distillation approach.
For the sake of the space, we report the results of BPR and LightGCN on CiteULike dataset with $\phi=0.1$ (Figure \ref{fig:hp}).
\textbf{1)} The batch size is an important factor affecting the performance of the topology distillation.
When the batch size is too small, the topology cannot include the overall relational knowledge in the representation space, leading to limited performance.
For CiteULike, we observe that \proposed achieves the stable performance around $2^{8}$-$2^{11}$.
In this work, we set the batch size to $2^{10}$.
\textbf{2)} $\gamma$ is a hyperparameter for balancing the topology-preserving loss and the regression loss.
We observe the stable performance with a value larger than $0.1$.
In this work, we set $\gamma$ to $0.5$.
Note that $\gamma=0$ equals DE.
\textbf{3)} The number of preference groups ($K$) is an important hyperparameter of \proposed. 
It needs to be determined considering the dataset, the capacity gap, and the selected layer for the distillation. 
For this setup, the best performance is achieved when $K$ is around 30-40 in both base models.
\textbf{4)} $\lambda_{TD}$ is a hyperparameter for controlling the effects of topology distillation.
For this setup, the best performance is achieved when $\lambda_{TD}$ is around $10^{-3}$ in both base models.

\section{Conclusion and Future Work}
\label{sec:conclusion}
We develop a general topology distillation approach for RS, which guides the learning of the student by the topological structure built upon the relational knowledge in the teacher representation space.
Concretely, we propose two topology distillation methods: 
1) FTD that transfers the full topology. 
FTD is used in the scenario where the student has enough capacity to learn all the teacher's knowledge.
2) \proposed that transfers the decomposed topology hierarchically.
\proposed is adopted in the conventional KD scenario where the student has a very limited capacity compared to the teacher.
We conduct extensive experiments on real-world datasets and show that the proposed approach consistently outperforms the state-of-the-art competitor. 
We also provide in-depth analyses to ascertain the benefit of distilling the topology.

We believe the topology distillation approach can be advanced and extended in several directions.
First, layer selection and simultaneous distillation from multiple layers are not investigated in this work.
We especially expect that this can further improve the limited improvements by the topology distillation in the deep model (i.e., NeuMF).
Second, topology distillation across different base models (e.g., from LightGCN to BPR) can be also considered to further improve the performance.
Lastly, prior knowledge of user/item groups (e.g., item category, user demographic features) can be utilized for more sophisticated topology decomposition.
We expect that this can further improve the effectiveness of the proposed method by providing better supervision on relational knowledge.

\vspace{0.3cm}
\noindent
\textbf{Acknowledgment:} This work was supported by the NRF grant funded by the MSIT (No. 2020R1A2B5B03097210), and the IITP grant funded by the MSIT (No. 2018-0-00584, 2019-0-01906).

\bibliographystyle{ACM-Reference-Format}
\balance
\bibliography{acmart}

\pagebreak
\newpage
\label{sec:appendix}
\appendix
\clearpage

\section{Appendix}

\subsection{Pseudocode of the proposed methods}
The pseudocode of FTD and \proposed are provided in Algorithm 1 and Algorithm 2, respectively.
The base model can be any existing recommender and $\mathcal{L}_{Base}$ is its loss function.
Note that the distillation is conducted in the offline training phase.
At the online inference phase, the student model is used only.

All the computations of the topology distillation are efficiently computed on matrix form by parallel execution through GPU processor.
In Algorithm 1 (FTD), the topological structures (line 5) are computed as follows: 
$$\mathbf{A}^{t} = \text{Cos}(\mathbf{E}^{t}, \mathbf{E}^{t})\;\;\text{and}\;\;\mathbf{A}^{s} = \text{Cos}(\mathbf{E}^{s}, \mathbf{E}^{s}),$$
where $\text{Cos}$ is the operation computing the cosine similarities by $\text{Cos}(\mathbf{B}, \mathbf{D}) = \hat{\mathbf{B}}\hat{\mathbf{D}}^{\top}$, $\hat{\mathbf{B}}_{[i,:]} = \mathbf{B}_{[i,:]} / \lVert \mathbf{B}_{[i,:]} \rVert_2$.
In Algorithm 2 (\proposed with Group(P,e)), the two-level topological structures (line 5-8) are computed as follows:
\begin{equation*}
\begin{aligned}
\mathbf{Z} = \text{Gumbel}&\text{-Softmax}(v(\mathbf{E}^t))\\
\mathbf{P}^{t} = \Tilde{\mathbf{Z}}^{\top}\mathbf{E}^{t} \;\;&\text{and}\;\; \mathbf{P}^{s} = \Tilde{\mathbf{Z}}^{\top}\mathbf{E}^{s}\\
\mathbf{H}^{t} = \text{Cos}(\mathbf{P}^{t}, \mathbf{E}^{t}) \;\;&\text{and}\;\; \mathbf{H}^{s} = \text{Cos}(\mathbf{P}^{s}, \mathbf{E}^{s})\\
\mathbf{A}^{t} = \text{Cos}(\mathbf{E}^{t}, \mathbf{E}^{t}) \;\;&\text{and}\;\; \mathbf{A}^{s} = \text{Cos}(\mathbf{E}^{s}, \mathbf{E}^{s})\\
\mathbf{M} &= \mathbf{Z} \mathbf{Z}^{\top}\\
\end{aligned}
\end{equation*}
The power of topology distillation comes from distilling the additional supervision in a proper manner considering the capacity gap.

\subsection{Group Assignment.}
Instead of DE, any clustering method or prior knowledge of user/item groups can be utilized for more sophisticated topology decomposition.
The simplest method is $K$-means clustering. 
Specifically, we first conduct the clustering in the teacher space, then use the results for the group assignment.
The results are summarized in Table \ref{tab:fixed}.
For both DE and $K$-means, the number of preference group ($K$) is set to 30, and the teacher space dimensions ($d^t$)~is~200.

We get consistently better results with the adaptive assignment by DE. 
We conjecture the possible reasons as follows:
1) Accurate clustering in high-dimensional space is very challenging,
2) With the adaptive approach, the assignment process gets gradually sophisticated along with the student, and thereby it provides guidance considering the student's learning.
For these reasons, we use the adaptive assignment in \proposed.
The performance of \proposed can be further improved by adopting a more advanced assignment method and prior knowledge.
We leave exploring better ways of the assignment for future study.

\begin{table}[h]
\centering
\caption{Performance comparison with different group assignment methods.}
\renewcommand{\tabcolsep}{1.8mm}
  \begin{minipage}[t]{1\linewidth}
  \centering
  \begin{tabular}{l cc cc}
    \toprule[.15em]
        &  \multicolumn{2}{c}{CiteULike}& \multicolumn{2}{c}{Foursquare}\\
    \cmidrule(lr){2-3}\cmidrule(lr){4-5}
     &Recall@50 & NDCG@50 & Recall@50 & NDCG@50 \\
    \midrule[.15em]
    $K$-means &0.2661&0.0980&0.2346&0.0871\\
    DE & 0.2803& 0.1031& 0.2438& 0.0921\\
    \bottomrule[.15em]
  \end{tabular}
  \end{minipage}
  \label{tab:fixed}
\end{table}

\vfill\null

\vspace{-10pt}
\begin{algorithm}[h]
\SetKwInOut{Input}{Input}
\SetKwInOut{Output}{Output}
\Input{Training data $\mathcal{D}$, Trained Teacher model}
\Output{Student model}
\BlankLine
Initialize Student model\\
\While {not convergence}{
\For{each batch $\mathcal{B} \in \mathcal{D}$}{
\BlankLine
Compute $\mathcal{L}_{Base}$ \\
\BlankLine
\tcc{Topology Distillation}
Build full topology $\mathbf{A}^t$ and $\mathbf{A}^s$\\
Compute $\mathcal{L}_{FTD}$ \COMMENT{\textit{Eqn. 5}}\\
\BlankLine
Compute $\mathcal{L} = \mathcal{L}_{Base} + \lambda \mathcal{L}_{FTD}$ \COMMENT{\textit{Eqn. 3}}\\
Update Student model by minimizing $\mathcal{L}$
}
}
\caption{Full Topology Distillation.}
\end{algorithm}
\vspace{-10pt}
\begin{algorithm}[h]
\SetKwInOut{Input}{Input}
\SetKwInOut{Output}{Output}
\Input{Training data $\mathcal{D}$, Trained Teacher model}
\Output{Student model}
\BlankLine
Initialize Student model, $v$, and $f_*$\\
\While {not convergence}{
\For{each batch $\mathcal{B} \in \mathcal{D}$}{
\BlankLine
Compute $\mathcal{L}_{Base}$ \\
\BlankLine
\tcc{Topology Distillation}
Assign preference groups $\mathbf{Z}$ for $\mathcal{B}$\\
Compute prototypes $\mathbf{P}^t$ and $\mathbf{P}^s$ \\
Build group-level topology $\mathbf{H}^t$ and $\mathbf{H}^s$\\
Build entity-level topology by $\mathbf{A}^t$, $\mathbf{A}^s$, and $\mathbf{M}$\\
Compute $\mathcal{L}_{HTD}$ \COMMENT{\textit{Eqn. 12}}\\
\BlankLine
Compute $\mathcal{L} = \mathcal{L}_{Base} + \lambda \mathcal{L}_{HTD}$ \COMMENT{\textit{Eqn. 3}}\\
Update Student model, $v$, and $f_*$ by minimizing $\mathcal{L}$
}
}
\caption{Hierarchical Topology Distillation.}
\end{algorithm}

\subsection{Datasets.}
We use two public real-world datasets: CiteULike and Foursquare.
We remove users having fewer than 5 (CiteULike) and 20 interactions (FourSquare) and remove items having fewer than 10 interactions (FourSquare) as done in \cite{DERRD, BUIR}.
Table \ref{tbl:statistic} summarizes the statistics of the datasets.
In the case of CiteULike, each item corresponds to an article, and each article has multiple tags.
In the case of Foursquare, each item corresponds to a POI (points-of-interest) such as museums and restaurants, and each POI has GPS coordinates (i.e., the latitude and longitude).
We use this side information in Section 4.3.
Table \ref{tbl:url} shows the URLs from which the datasets can be downloaded.

\begin{table}[h]
\centering
\renewcommand{\tabcolsep}{2.6mm}
  \caption{Statistics of the datasets.}
  \begin{tabular}{ccccc}
    \toprule[.15em]
    Dataset & \#Users & \#Items & \#Interactions & Density \\
    \midrule[.15em]
    CiteULike & 5,220 & 25,182 & 115,142 & 0.09\% \\
    Foursquare & 19,466 & 28,594 & 609,655 & 0.11\% \\
    \bottomrule[.15em]
  \end{tabular}
    \label{tbl:statistic}
    \vspace{-0.4cm}
\end{table}

\begin{table}[h]
\centering
  \caption{URL links to the datasets.}
  \begin{tabular}{c l}
    \toprule[.15em]
    Dataset & URL link to the dataset \\
    \midrule[.15em]
    CiteULike & \url{https://github.com/changun/CollMetric} \\
    Foursquare & \url{http://spatialkeyword.sce.ntu.edu.sg/eval-vldb17/} \\
    \bottomrule[.15em]
  \end{tabular}
    \label{tbl:url}
\end{table}

\subsection{Evaluation Protocol and Metrics}
We adopt the widely used \textit{leave-one-out} evaluation protocol, whereby two interacted items for each user are held out
for testing/validation, and the rest are used for training.
However, unlike \cite{DERRD} that samples a predefined number (e.g., 499) of unobserved items for evaluation, we adopt the full-ranking evaluation scheme
that evaluates how well each method can rank the test item higher than all the unobserved items.
Although it is time-consuming, it enables a more thorough evaluation compared to the sampling-based evaluation \cite{CD, krichene2020sampled}.
We evaluate all methods by two widely used ranking metrics: 
Recall@$N$ \cite{recall} and Normalized Discounted Cumulative Gain (NDCG@$N$) \cite{jarvelin2002cumulated}.
Recall@$N$ measures whether the test item is included in the top-$N$ list and NDCG@$N$ assigns higher scores on the upper ranked test items.
We compute the metrics for each user, then compute the average score.
Lastly, we report the average value of five independent runs for all methods.

\subsection{Implementation Details.}
We use PyTorch to implement the proposed methods and all the competing methods.
We optimize all methods with Adam optimizer.
For DE and RRD, we use the public implementation provided by the authors.
For each setting, hyperparameters are tuned by using grid searches on the validation set.
The learning rate is searched in the range of \{0.01, 0.005, 0.001, 0.0005, 0.0001\}.
The model regularizer is searched in the range of $\{10^{-2}, 10^{-3}, 10^{-4}, 10^{-5}, 10^{-6}\}$.
We set the total number of epochs to 500 and adopt the early stopping strategy; it terminates when Recall@50 on the validation set does not increase for 20 successive epochs.

For all base models, the number of negative samples is set to 1.
For NeuMF and LightGCN, the number of layers is searched in the range of \{1, 2, 3, 4\}.
For all the distillation methods, weight for the distillation loss ($\lambda$) searched in the range of $\{1, 10^{-1}, 10^{-2}, 10^{-3}, 10^{-4}, 10^{-5}\}$.
For the hint regression-related setup, we closely follow the setup reported in DE paper \cite{DERRD}.
Specifically, two-layer MLP with [$d^s \rightarrow (d^s + d^t)/2 \rightarrow d^t$] is employed for $f$ in FitNet, DE and \proposed.
Also, one-layer perceptron with [$d^t \rightarrow K $] is employed for assigning group ($v$) in DE and \proposed.
For DE and \proposed, the number of preference groups ($K$) is chosen from \{5, 10, 20, 30, 40, 50\}.
We provide an analysis of these hyperparameters in Section 4.4.

\subsection{Experiment Setup for Downstream Tasks.}
We evaluate how well each method encodes the items’ characteristics (or semantics) into the representations.
We train a small network to predict the side information of items by using the \textit{fixed} item representations as the input.
Specifically, we use a linear and a non-linear model (i.e., a single-layer perceptron and three-layer perceptron, respectively) with Softmax output.
The linear model has the shape of [$d^s \rightarrow C$], and the non-linear model has the shape of [$d^s \rightarrow (d^s+C)/2 \rightarrow (d^s+C)/2 \rightarrow C$] with relu, where $C$ is the number of tags/classes.
Let $\mathbf{q}$ denote the output of the model whose element $q_i$ is a prediction score for each tag/class.
Also, let $\mathbf{p}$ denote the ground-truth vector whose element $p_i=1$ if $i$-th tag/class is the answer, otherwise $p_i=0$. 
We train the model by minimizing the negative log-likelihood: $-\sum_{i} p_i \log q_i$.
Note that the side-information is not utilized for training of the base model.

For CiteULike dataset, we perform \textbf{item-tag retrieval task};
by using each item representation as a query, we find a ranking list of tags that are relevant to the item.
We first remove tags used less than 10 times.
Then, there exist 4,153 tags and an item has 6.4 tags on average.
After training, we make a ranking list of tags by sorting the prediction scores.
We evaluate how many the ground-truth tags are included in the top-$10$ list by Recall$@10$.
For Foursquare dataset, we perform \textbf{item-region classification task};
given each item representation, we predict the region class to which the item belongs.
We first perform $k$-means clustering on the coordinates with $k=200$ and use the clustering results as the class labels. 
After training, we evaluate the performance by Accuracy.
Finally, we perform 5-fold cross-validation and report the average result and standard deviation in Table \ref{tbl:downstream_tasks}.

\vspace{20cm}

\end{document}